%% file: paper.tex
\documentclass[11pt,a4paper]{article}
\usepackage[hyperref]{acl2019}
\usepackage{times}
\usepackage{latexsym}

\usepackage{url}

\aclfinalcopy 
\def\aclpaperid{2220} 

\usepackage{adjustbox}
\usepackage{amsmath}
\usepackage{booktabs}
\usepackage{multirow}
\usepackage{supertabular}
\usepackage{placeins}
\usepackage{enumitem}
\graphicspath{{./fig/}{./fig_raw/}}

\DeclareMathOperator{\median}{median}
\DeclareMathOperator*{\argmin}{argmin}

\makeatletter
\newcommand\Biggg{\bBigg@{3.5}}
\makeatother
\newcommand\secref[1]{\S\ref{#1}}

\title{%
	Sequence Tagging with Contextual and Non-Contextual \\
	Subword Representations: A Multilingual Evaluation}

\author{Benjamin Heinzerling\textsuperscript{\textnormal{$\dag$}}\thanks{\hspace{0.4em} Work done while at HITS.} \hspace{0.18em} \textnormal{and}\hspace{0.09em} Michael Strube\textsuperscript{\textnormal{$\ddag$}}\\
\vspace{0.1ex}
  \textsuperscript{\textnormal{$\dag$}}RIKEN AIP \& Tohoku University \\
\vspace{0.1ex}
  \textsuperscript{\textnormal{$\ddag$}}Heidelberg Institute for Theoretical Studies gGmbH \\
\vspace{0.1ex}
  \hypersetup{urlcolor=black}\href{mailto:benjamin.heinzerling@riken.jp}{\tt benjamin.heinzerling@riken.jp} \hspace{0.06em} $\vert$ \hspace{0.12em} \hypersetup{urlcolor=black}\href{mailto:michael.strube@h-its.org}{\tt michael.strube@h-its.org} \\
}

\date{}

\begin{document}
\maketitle
\begin{abstract}
Pretrained contextual and non-contextual subword embeddings have become available in over 250 languages, allowing massively multilingual NLP.
However, while there is no dearth of pretrained embeddings, the distinct lack of systematic evaluations makes it difficult for practitioners to choose between them.
In this work, we conduct an extensive evaluation comparing non-contextual subword embeddings, namely FastText and BPEmb, and a contextual representation method, namely BERT, on multilingual named entity recognition and part-of-speech tagging.

We find that overall, a combination of BERT, BPEmb, and character representations works well across languages and tasks.
A more detailed analysis reveals different strengths and weaknesses: Multilingual BERT performs well in medium- to high-resource languages, but is outperformed by non-contextual subword embeddings in a low-resource setting.

\end{abstract}

\section{Introduction}

Rare and unknown words pose a difficult challenge for embedding methods that rely on seeing a word frequently during training \citep{bullinaria2007extracting,luong2013word}.
Subword segmentation methods avoid this problem by assuming a word's meaning can be inferred from the meaning of its parts.
Linguistically motivated subword approaches first split words into morphemes and then represent word meaning by composing morpheme embeddings \citep{luong2013word}.
More recently, character-ngram approaches \citep{luong2016open,bojanowski2017subword} and Byte Pair Encoding (BPE) \citep{sennrich2016subword} have grown in popularity, likely due to their computational simplicity and language-agnosticity.\footnote{%
	While language-agnostic, these approaches are not language-independent. See Appendix~\ref{app:language-independence} for a discussion.}

\begin{figure}[t!]
	\centering
	\includegraphics[width=\linewidth]{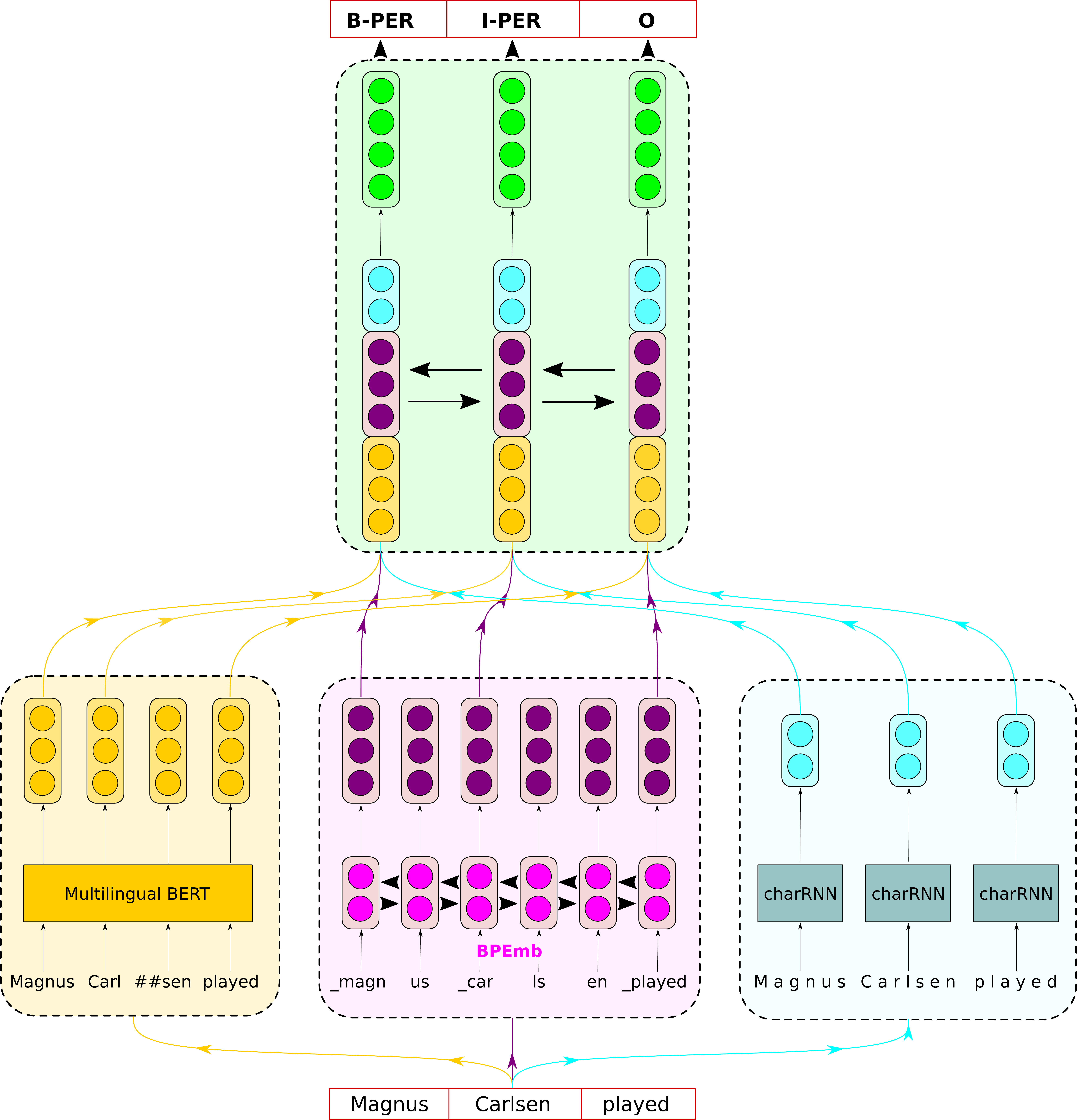}
	\caption{A high-performing ensemble of subword representations encodes the input using multilingual BERT (yellow, bottom left), an LSTM with BPEmb (pink, bottom middle), and a character-RNN (blue, bottom right). A meta-LSTM (green, center) combines the different encodings before classification (top). Horizontal arrows symbolize bidirectional LSTMs.}
	\label{fig:model}
\end{figure}

\begin{table*}[!htbp]
	\centering
	\small
	\begin{adjustbox}{max width=\linewidth}
		\begin{tabular}{l|llllll}
			\toprule
			Method & \multicolumn{6}{c}{Subword segmentation and token transformation}\\
			\midrule
			Original text & Magnus & Carlsen & played & against & Viswanathan & Anand \\
			\rule{0pt}{10pt}%
			Characters & M a g n u s & C a r l s e n & p l a y e d & a g a i n s t & V i s w a n a t h a n & A n a n d \\
			Word shape & Aa & Aa & a & a & Aa & Aa \\
			\rule{0pt}{10pt}%
			FastText & magnus+mag+\ldots & carlsen+car+arl+\ldots & played+\ldots & against+\ldots & vis+isw+\ldots+nathan & ana+\ldots \\
			\rule{0pt}{10pt}%
			BPE vs1000 & \_m ag n us & \_car l s en & \_play ed & \_against & \_v is w an ath an & \_an and \\
			BPE vs3000 & \_mag n us & \_car ls en & \_played & \_against & \_vis w an ath an & \_an and \\
			BPE vs5000 & \_magn us & \_car ls en & \_played & \_against & \_vis wan ath an & \_an and \\
			BPE vs10000 & \_magn us & \_car ls en & \_played & \_against & \_vis wan athan & \_an and \\
			BPE vs25000 & \_magnus & \_car ls en & \_played & \_against & \_vis wan athan & \_an and \\
			BPE vs50000 & \_magnus & \_carls en & \_played & \_against & \_vis wan athan & \_anand \\
			BPE vs100000 & \_magnus & \_carlsen & \_played & \_against & \_viswan athan & \_anand \\
			\rule{0pt}{10pt}%
			BERT & Magnus & Carl \#\#sen & played & against & V \#\#is \#\#wana \#\#than & Anand \\
			\bottomrule
		\end{tabular}
	\end{adjustbox}
	\caption{Overview of the subword segmentations and token transformations evaluated in this work.}
	\label{tbl:subword-examples}
\end{table*}

\noindent\textbf{Sequence tagging with subwords.} Subword information has long been recognized as an important feature in sequence tagging tasks such as named entity recognition (NER) and part-of-speech (POS) tagging.
For example, the suffix \emph{-ly} often indicates adverbs in English POS tagging and English NER may exploit that professions often end in suffixes like \emph{-ist} (\emph{journalist}, \emph{cyclist}) or companies in suffixes like \emph{-tech} or \emph{-soft}.
In early systems, these observations were operationalized with manually compiled lists of such word endings or with character-ngram features \citep{nadeau2007survey}.
Since the advent of neural sequence tagging \citep{graves2012supervised,huang2015bidirectional}, the predominant way of incorporating character-level subword information is learning embeddings for each character in a word, which are then composed into a fixed-size representation using a character-CNN \citep{chiu2016named} or character-RNN (char-RNN) \citep{lample2016neural}.
Moving beyond single characters, pretrained subword representations such as FastText, BPEmb, and those provided by BERT (see §\ref{sec:subword-embeddings}) have become available.

While there now exist several pretrained subword representations in many languages, a practitioner faced with these options has a simple question: Which subword embeddings should I use?
In this work, we answer this question for multilingual named entity recognition and part-of-speech tagging and make the following contributions:%
\begin{itemize}
	\setlength{\itemsep}{-3pt}
	\setlength{\parsep}{1pt}
	\item We present a large-scale evaluation of multilingual subword representations on two sequence tagging tasks;
	\item We find that subword vocabulary size matters and give recommendations for choosing it;
	\item We find that different methods have different strengths: Monolingual BPEmb works best in medium- and high-resource settings, multilingual non-contextual subword embeddings are best in low-resource languages, while multilingual BERT gives good or best results across languages.
\end{itemize}

\section{Subword Embeddings}
\label{sec:subword-embeddings}

We now introduce the three kinds of multilingual subword embeddings compared in our evaluation: FastText and BPEmb are collections of pretrained, monolingual, non-contextual subword embeddings available in many languages, while BERT provides contextual subword embeddings for many languages in a single pretrained language model with a vocabulary shared among all languages.
Table~\ref{tbl:subword-examples} shows examples of the subword segmentations these methods produce.

\subsection{FastText: Character-ngram Embeddings}

FastText \citep{bojanowski2017subword} represents a word $w$ as the sum of the learned embeddings $\vec{z}_g$ of its constituting character-ngrams $g$ and, in case of in-vocabulary words, an embedding $\vec{z_w}$ of the word itself: $\vec{w} = \vec{z_w} + \sum_{g \in G_w} \vec{z}_g$, where $G_w$ is the set of all constituting character n-grams for $3 \le n \le 6$.
\citeauthor{bojanowski2017subword} provide embeddings trained on Wikipedia editions in 294 languages.\footnote{%
	\url{https://fasttext.cc/docs/en/pretrained-vectors.html}}

\subsection{BPEmb: Byte-Pair Embeddings}
Byte Pair Encoding (BPE) is an unsupervised segmentation method which operates by iteratively merging frequent pairs of adjacent symbols into new symbols.
E.g., when applied to English text, BPE merges the characters \emph{h} and \emph{e} into the new byte-pair symbol \emph{he}, then the pair consisting of the character \emph{t} and the byte-pair symbol \emph{he} into the new symbol \emph{the} and so on.
These merge operations are learned from a large background corpus.
The set of byte-pair symbols learned in this fashion is called the \emph{BPE vocabulary}.

Applying BPE, i.e.\,iteratively performing learned merge operations, segments a text into subwords (see BPE segmentations for vocabulary sizes \emph{vs1000} to \emph{vs100000} in Table~\ref{tbl:subword-examples}).
By employing an embedding algorithm, e.g.\,GloVe \citep{pennington2014glove}, to train embeddings on such a subword-segmented text, one obtains embeddings for all byte-pair symbols in the BPE vocabulary.
In this work, we evaluate BPEmb \citep{heinzerling2018bpemb}, a collection of byte-pair embeddings trained on Wikipedia editions in 275 languages.\footnote{%
	\url{https://nlp.h-its.org/bpemb/}}

\subsection{BERT: Contextual Subword Embeddings}

One of the drawbacks of the subword embeddings introduced above, and of pretrained word embeddings in general, is their lack of context.
For example, with a non-contextual representation, the embedding of the word \emph{play} will be the same both in the phrase \emph{a play by Shakespeare} and the phrase \emph{to play Chess}, even though \emph{play} in the first phrase is a noun with a distinctly different meaning than the verb \emph{play} in the second phrase.
Contextual word representations \citep{dai2015semi,melamud2016context2vec,ramachandran2017unsupervised,peters2018contextualized,radford2018improving,howard2018universal} overcome this shortcoming via pretrained language models.

Instead of representing a word or subword by a lookup of a learned embedding, which is the same regardless of context, a contextual representation is obtained by encoding the word in context using a neural language model \citep{bengio2003neural}.
Neural language models typically employ a sequence encoder such as a bidirectional LSTM \citep{hochreiter1997long} or Transformer \citep{vaswani2017attention}.
In such a model, each word or subword in the input sequence is encoded into a vector representation.
With a bidirectional LSTM, this representation is influenced by its left and right context through state updates when encoding the sequence from left to right and from right to left.
With a Transformer, context influences a word's or subword's representation via an attention mechanism \citep{bahdanau2014neural}.

In this work we evaluate BERT \citep{devlin2018bert}, a Transformer-based pretrained language model operating on subwords similar to BPE (see last row in Table~\ref{tbl:subword-examples}).
We choose BERT among the pretrained language models mentioned above since it is the only one for which a multilingual version is publicly available.
Multilingual BERT\footnote{%
	\url{https://github.com/google-research/bert/blob/f39e881/multilingual.md}}
has been trained on the 104 largest Wikipedia editions, so that, in contrast to FastText and BPEmb, many low-resource languages are not supported.

\section{Multilingual Evaluation}
\label{sec:eval}
\begin{table}
	\centering
	\small
	\begin{adjustbox}{max width=\linewidth}
		\begin{tabular}{lrll}
			\toprule
			Method & \#languages & Intersect. 1 & Intersect. 2\\
			\midrule
			FastText & 294 & \multirow{3}{2em}{\Bigg{\}}265} & \multirow{4}{2em}{\Biggg{\}}101} \\ 
			Pan17 & 282 & \\
			BPEmb & 275 &  \\
			BERT & 104  & - \\
			\bottomrule
		\end{tabular}
	\end{adjustbox}
	\caption{Number of languages supported by the three subword embedding methods compared in our evaluation, as well as the NER baseline system (Pan17).}
	\label{tbl:num-langs}
\end{table}
We compare the three different pretrained subword representations introduced in \secref{sec:subword-embeddings} on two tasks: NER and POS tagging.
Our multilingual evaluation is split in four parts.
After devising a sequence tagging architecture (\secref{sec:model}), we investigate an important hyper-parameter in BPE-based subword segmentation: the BPE vocabulary size (\secref{sec:tuning-bpe}).
Then, we conduct NER experiments on two sets of languages (see Table~\ref{tbl:num-langs}): 265 languages supported by FastText and BPEmb (\secref{sec:ner-all}) and the 101 languages supported by all methods including BERT (\secref{sec:ner-bert}).
Our experiments conclude with POS tagging on 27 languages (\secref{sec:pos-tagging}).

\noindent{\textbf{Data.}} For NER, we use WikiAnn \citep{pan2017crosslingual}, a dataset containing named entity mention and three-class entity type annotations in 282 languages.
WikiAnn was automatically generated by extracting and classifying entity mentions from inter-article links on Wikipedia.
Because of this, WikiAnn suffers from problems such as skewed entity type distributions in languages with small Wikipedias (see Figure~\ref{fig:tag-distributions} in Appendix~\ref{sec:analysis-wikiann}), as well as wrong entity types due to automatic type classification.
These issues notwithstanding, WikiAnn is the only available NER dataset that covers almost all languages supported by the subword representations compared in this work.
For POS tagging, we follow \citet{plank2016multilingual,yasunaga2018robust} and use annotations from the Universal Dependencies project \citep{nivre2016universal}.
These annotations take the form of language-universal POS tags \citep{petrov2012universal}, such as \emph{noun}, \emph{verb}, \emph{adjective}, \emph{determiner}, and \emph{numeral}.

\subsection{Sequence Tagging Architecture}
\label{sec:model}
Our sequence tagging architecture is depicted in Figure~\ref{fig:model}.
The architecture is modular and allows encoding text using one or more subword embedding methods.
The model receives a sequence of tokens as input, here \emph{Magnus Carlsen played}.
After subword segmentation and an embedding lookup, subword embeddings are encoded with an encoder specific to the respective subword method.
For BERT, this is a pretrained Transformer, which is finetuned during training.
For all other methods we train bidirectional LSTMs.
Depending on the particular subword method, input tokens are segmented into different subwords.
Here, BERT splits \emph{Carlsen} into two subwords resulting in two encoder states for this token, while BPEmb with an LSTM encoder splits this word into three.
FastText (not depicted) and character RNNs yield one encoder state per token.
To match subword representations with the tokenization of the gold data, we arbitrarily select the encoder state corresponding to the first subword in each token.
A meta-LSTM combines the token representations produced by each encoder before classification.\footnote{%
	In preliminary experiments (results not shown), we found that performing classification directly on the concatenated token representation without such an additional LSTM on top does not work well.}

Decoding the sequence of a neural model's pre-classification states with a conditional random field (CRF) \citep{lafferty01} has been shown to improve NER performance by 0.7 to 1.8 F1 points \citep{ma2016endtoend,reimers2017distributions} on a benchmark dataset.
In our preliminary experiments on WikiAnn, CRFs considerably increased training time but did not show consistent improvements across languages.\footnote{%
	The system we compare to as baseline \citep{pan2017crosslingual} includes a CRF but did not report an ablation without it.}
Since our study involves a large number of experiments comparing several subword representations with cross-validation in over 250 languages, we omit the CRF in order to reduce model training time.

\noindent\textbf{Implementation details.} Our sequence tagging architecture is implemented in PyTorch \citep{paszke2017pytorch}.
All model hyper-parameters for a given subword representation are tuned in preliminary experiments on development sets and then kept the same for all languages (see Appendix~\ref{sec:hyper-parameters}).
For many low-resource languages, WikiAnn provides only a few hundred instances with skewed entity type distributions.
In order to mitigate the impact of variance from random train-dev-test splits in such cases, we report averages of n-fold cross-validation runs, with n=10 for low-resource, n=5 for medium-resource, and n=3 for high-resource languages.\footnote{%
	Due to high computational resource requirements, we set n=1 for finetuning experiments with BERT.}
For experiments involving FastText, we precompute a 300d embedding for each word and update embeddings during training.
We use BERT in a \emph{finetuning} setting, that is, we start training with a pretrained model and then update that model's weights by backpropagating through all of BERT's layers.
Finetuning is computationally more expensive, but gives better results than feature extraction, i.e.\ using one or more of BERT's layers for classification without finetuning \citep{devlin2018bert}.
For BPEmb, we use 100d embeddings and choose the best BPE vocabulary size as described in the next subsection.

\subsection{Tuning BPE}
\label{sec:tuning-bpe}
\begin{figure}
	\centering
	\includegraphics[width=\linewidth]{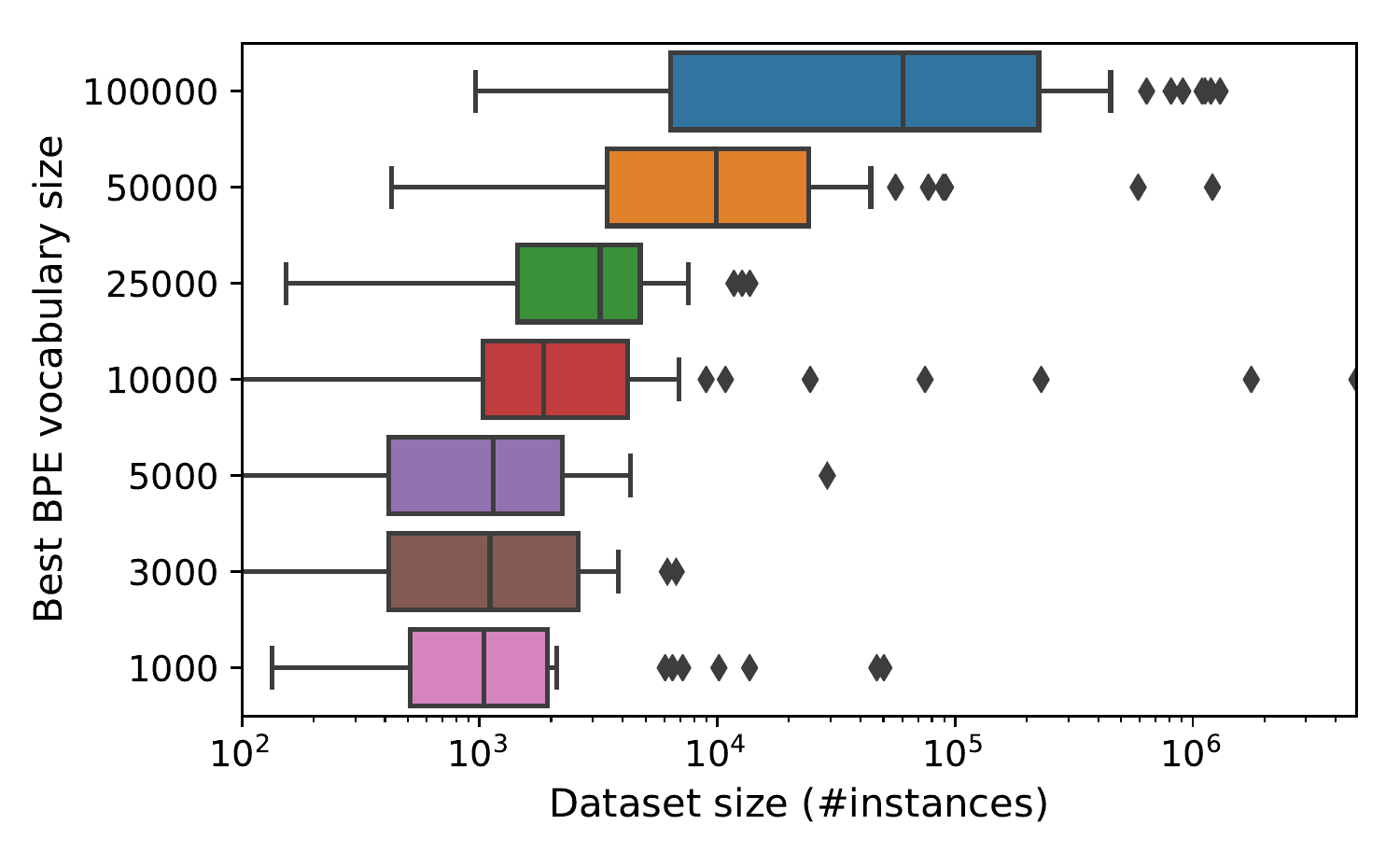}
	\caption{The best BPE vocabulary size varies with dataset size. For each of the different vocabulary sizes, the box plot shows means and quartiles of the dataset sizes for which this vocabulary size is optimal, according to the NER F1 score on the respective development set in WikiAnn. E.g., the bottom, pink box records the sizes of the datasets (languages) for which BPE vocabulary size 1000 was best, and the top, blue box the dataset sizes for which vocabulary size 100k was best.}
	\label{fig:best-vocabsizes}
\end{figure}

In subword segmentation with BPE, performing only a small number of byte-pair merge operations results in a small vocabulary.
This leads to oversegmentation, i.e., words are split into many short subwords (see \emph{BPE vs1000} in Table~\ref{tbl:subword-examples}).
With more merge operations, both the vocabulary size and the average subword length increase.
As the byte-pair vocabulary grows larger it adds symbols corresponding to frequent words, resulting in such words not being split into subwords.
Note, for example, that the common English preposition \emph{against} is not split even with the smallest vocabulary size, or that \emph{played} is split into the stem \emph{play} and suffix \emph{ed} with a vocabulary of size 1000, but is not split with larger vocabulary sizes.

\begin{table*}[t!]
	\centering
	\small
	\begin{adjustbox}{max width=\linewidth}
		\input{tables_raw/wikiann_overall_results_all_langs}
	\end{adjustbox}
	\caption{NER results on WikiAnn. The first row shows macro-averaged F1 scores (\%) for all 265 languages in the \emph{Intersect.\,1} setting. Rows two to four break down scores for 188 low-resource languages (\textless10k instances), 48 medium-resource languages (10k to 100k instances), and 29 high-resource languages (\textgreater100k instances).}
	\label{tbl:wikiann-results-all}
\end{table*}

The choice of vocabulary size involves a trade-off.
On the one hand, a small vocabulary requires less data for pre-training subword embeddings since there are fewer subwords for which embeddings need to be learned.
Furthermore, a smaller vocabulary size is more convenient for model training since training time increases with vocabulary size \citep{morin2005hierarchical} and hence a model with a smaller vocabulary trains faster.
On the other hand, a small vocabulary results in less meaningful subwords and longer input sequence lengths due to oversegmentation.

Conversely, a larger BPE vocabulary tends to yield longer, more meaningful subwords so that subword composition becomes easier -- or in case of frequent words even unnecessary -- in downstream applications, but a larger vocabulary also requires a larger text corpus for pre-training good embeddings for all symbols in the vocabulary.
Furthermore, a larger vocabulary size requires more annotated data for training larger neural models and increases training time.

Since the optimal BPE vocabulary size for a given dataset and a given language is not a priori clear, we determine this hyper-parameter empirically.
To do so, we train NER models with varying BPE vocabulary sizes\footnote{%
	We perform experiments with vocabulary sizes in $\{1000, 3000, 5000, 10000, 25000, 50000, 100000\}$.
	} for each language and record the best vocabulary size on the language's development set as a function of dataset size (Figure~\ref{fig:best-vocabsizes}).
This data shows that larger vocabulary sizes are better for high-resource languages with more training data, and smaller vocabulary sizes are better for low-resource languages with smaller datasets.
In all experiments involving byte-pair embeddings, we choose the BPE vocabulary size for the given language according to this data.\footnote{%
	The procedure for selecting BPE vocabulary size is given in Appendix~\ref{sec:best-vocab-size}.}

\subsection{NER with FastText and BPEmb}
\label{sec:ner-all}

In this section, we evaluate FastText and BPEmb on NER in 265 languages.
As baseline, we compare to \citet{pan2017crosslingual}'s system, which combines morphological features mined from Wikipedia markup with cross-lingual knowledge transfer via Wikipedia language links (\emph{Pan17} in Table~\ref{tbl:wikiann-results-all}).
Averaged over all languages, FastText performs 4.1 F1 points worse than this baseline.
BPEmb is on par overall, with higher scores for medium- and high-resource languages, but a worse F1 score on low-resource languages.
BPEmb combined with character embeddings (\emph{+char}) yields the overall highest scores for medium- and high-resource languages among monolingual methods.

\begin{figure}[t!]
	\centering
	\includegraphics[width=.91\linewidth]{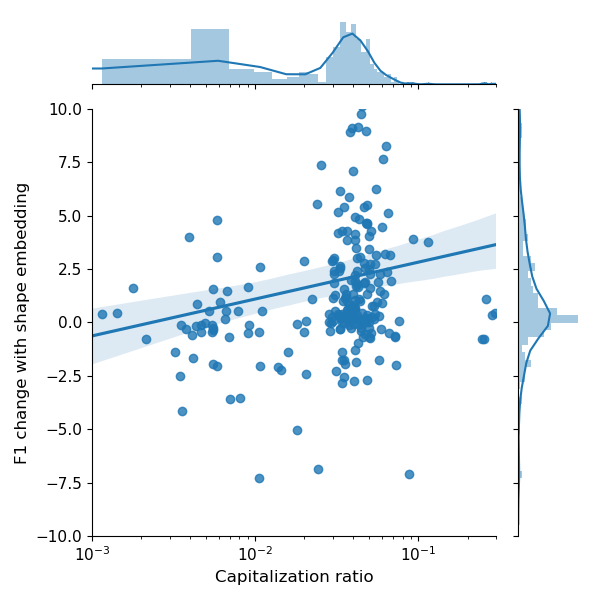}
	\caption{Impact of word shape embeddings on NER performance in a given language as function of the capitalization ratio in a random Wikipedia sample.}
	\label{fig:shape-benefit}
\end{figure}
\noindent\textbf{Word shape}. When training word embeddings, lowercasing is a common preprocessing step \citep{pennington2014glove} that on the one hand reduces vocabulary size, but on the other loses information in writing systems with a distinction between upper and lower case letters.
As a more expressive alternative to restoring case information via a binary feature indicating capitalized or lowercased words \cite{curran2003language}, word shapes \citep{collins2002ranking,finkel2005nonlocal} map characters to their type and collapse repeats.
For example, \emph{Magnus} is mapped to the word shape \emph{Aa} and \emph{G.M.} to \emph{A.A.}
Adding such shape embeddings to the model (\emph{+shape} in Table~\ref{tbl:wikiann-results-all}) yields similar improvements as character embeddings.

Since capitalization is not important in all languages, we heuristically decide whether shape embeddings should be added for a given language or not.
We define the \emph{capitalization ratio} of a language as the ratio of upper case characters among all characters in a written sample.
As Figure~\ref{fig:shape-benefit} shows, capitalization ratios vary between languages, with shape embeddings tending to be more beneficial in languages with higher ratios.
By thresholding on the capitalization ratio, we only add shape embeddings for languages with a high ratio (\emph{+someshape}).
This leads to an overall higher average F1 score of $85.3$ among monolingual models, due to improved performance ($81.9$ vs.\ $81.5$) on low-resource languages.

\begin{figure}
	\includegraphics[width=\linewidth]{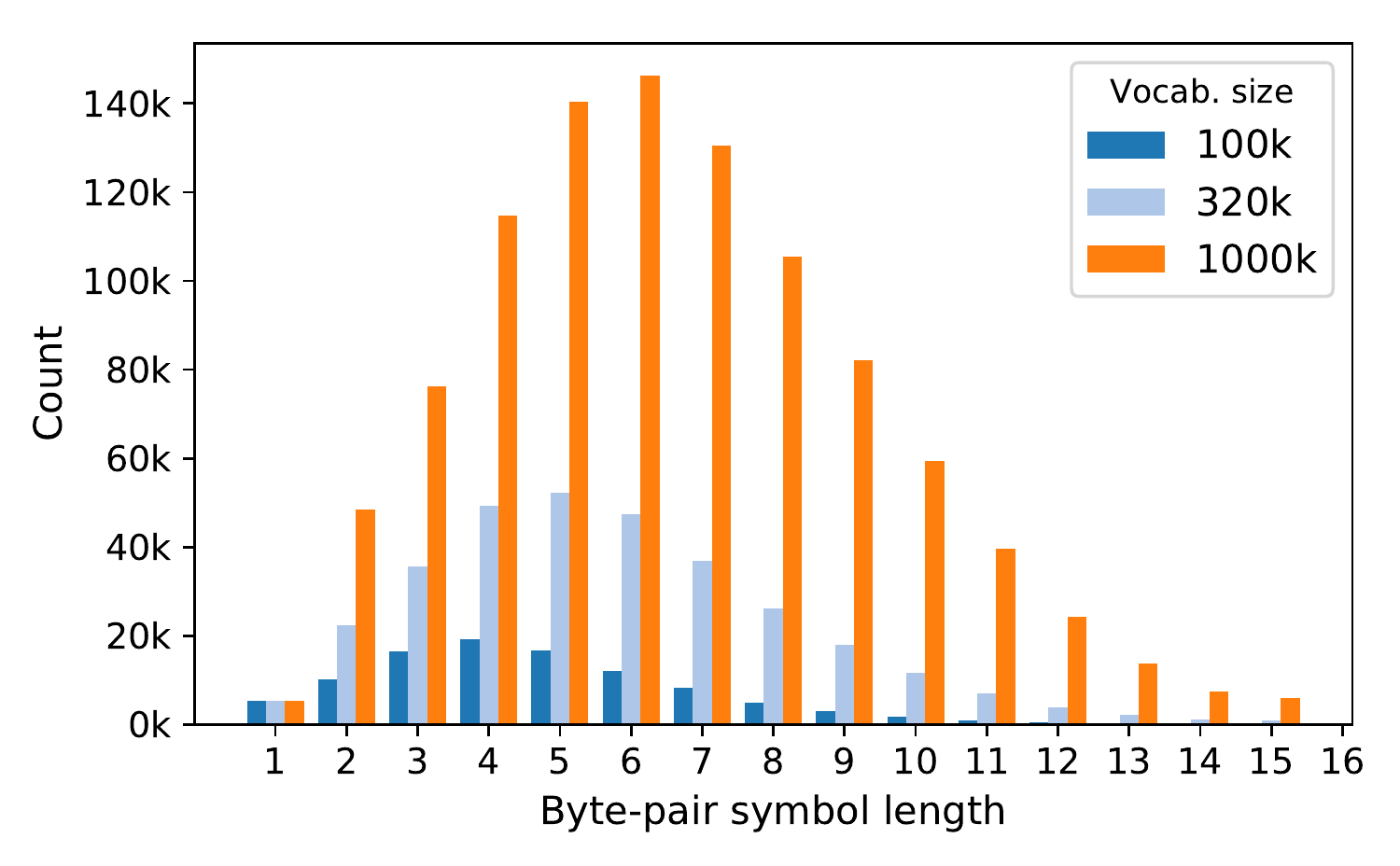}
	\vspace{-7mm}
	\caption{The distribution of byte-pair symbol lengths varies with BPE vocabulary size.}
	\label{fig:bpe-length}
\end{figure}

\begin{table}
	\centering
	\small
	\begin{tabular}{lrrr}
		\toprule
		 & \multicolumn{3}{c}{BPE vocabulary size} \\
		 & 100k & 320k & 1000k \\
		\midrule
		Dev. F1 & 87.1 & 88.7 & 89.3 \\
		\bottomrule
	\end{tabular}
	\caption{Average WikiAnn NER F1 scores on the development sets of 265 languages with shared vocabularies of different size.}
	\label{tbl:bpe-vocab-size}
\end{table}

\begin{figure*}
	\centering
	\includegraphics[width=0.41\linewidth,trim={1.9cm 1.5cm 1.5cm 1.7cm},clip]{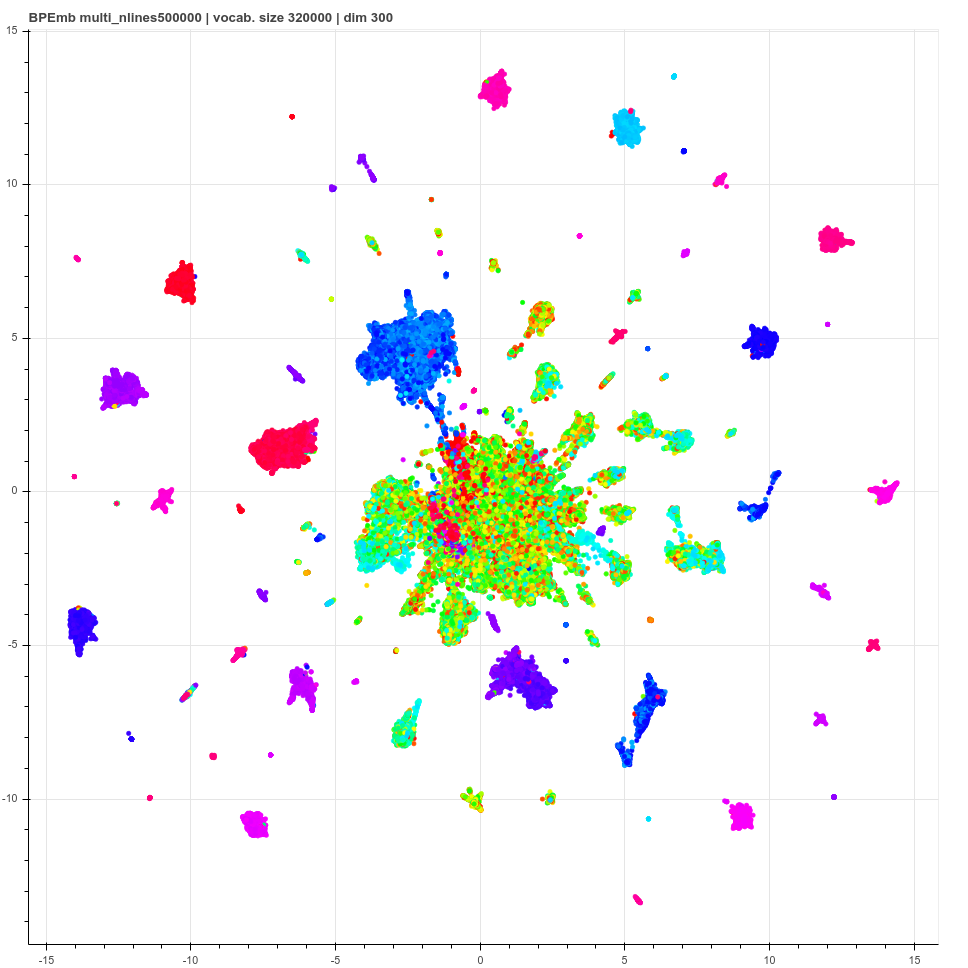}
	\qquad \qquad
	\includegraphics[width=0.41\linewidth,trim={1.9cm 1.5cm 1.5cm 1.6cm},clip]{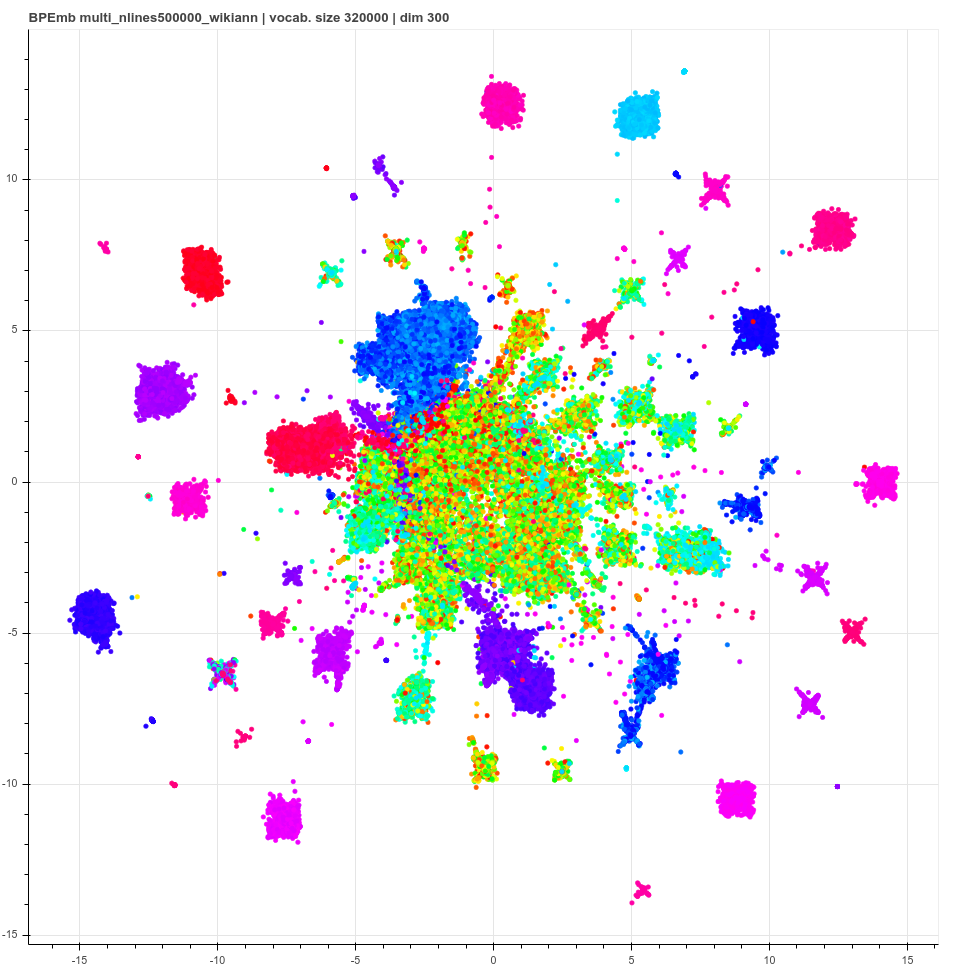}
	\vspace{-1.5ex}
	\caption{Shared multilingual byte-pair embedding space pretrained (left) and after NER model training (right), 2-d UMAP projection \citep{mcinnes2018umap}. As there is no 1-to-1 correspondence between BPE symbols and languages in a shared multilingual vocabulary, it is not possible to color BPE symbols by language.
Instead, we color symbols by Unicode code point.
	This yields a coloring in which, for example, BPE symbols consisting of characters from the Latin alphabet are green (large cluster in the center), symbols in Cyrillic script blue (large cluster at 11 o'clock), and symbols in Arabic script purple (cluster at 5 o'clock). Best viewed in color.}
	\label{fig:multi-emb-training}
\end{figure*}

\begin{table*}[t!]
	\centering
	\small
	\begin{adjustbox}{max width=\linewidth}
		\input{tables_raw/wikiann_overall_results_bert_langs}
	\end{adjustbox}
	\caption{NER F1 scores for the 101 WikiAnn languages supported by all evaluated methods.}
	\label{tbl:wikiann-results-bert}
\end{table*}

\noindent\textbf{One NER model for 265 languages.} The reduction in vocabulary size achieved by BPE is a crucial advantage in neural machine translation \citep{johnson2017googles} and other tasks which involve the costly operation of taking a softmax over the entire output vocabulary \citep[see][]{morin2005hierarchical,li2019efficient}.
BPE vocabulary sizes between 8k and 64k are common in neural machine translation.
Multilingual BERT operates on a subword vocabulary of size 100k which is shared among 104 languages. Even with shared symbols among languages, this allots at best only a few thousand byte-pair symbols to each language.
Given that sequence tagging does not involve taking a softmax over the vocabulary, much larger vocabulary sizes are feasible, and as \secref{sec:tuning-bpe} shows, a larger BPE vocabulary is better when enough training data is available.
To study the effect of a large BPE vocabulary size in a multilingual setting, we train BPE models and byte-pair embeddings with subword vocabularies of up to 1000k BPE symbols, which are shared among all languages in our evaluation.\footnote{%
	Specifically, we extract up to 500k randomly selected paragraphs from articles in each Wikipedia edition, yielding 16GB of text in 265 languages.
	Then, we train BPE models with vocabulary sizes 100k, 320k, and 1000k using SentencePiece \citep{kudo2018sentencepiece}, and finally train 300d subword embeddings using GloVe.}

The shared BPE vocabulary and corresponding byte-pair embeddings allow training a single NER model for all 265 languages.
To do so, we first encode WikiAnn in all languages using the shared BPE vocabulary and then train a single multilingual NER model in the same fashion as a monolingual model.
As the vocabulary size has a large effect on the distribution of BPE symbol lengths (Figure~\ref{fig:bpe-length}, also see \secref{sec:tuning-bpe}) and model quality, we determine this hyper-parameter empirically (Table~\ref{tbl:bpe-vocab-size}).
To reduce the disparity between dataset sizes of different languages, and to keep training time short, we limit training data to a maximum of 3000 instances per language.\footnote{%
With this limit, training takes about a week on one NVIDIA P40 GPU.}
Results for this multilingual model (\emph{MultiBPEmb}) with shared character embeddings (\emph{+char}) and without further finetuning \emph{-finetune} show a strong improvement in low-resource languages (89.7 vs.\ 81.9 with \emph{+someshape}), while performance degrades drastically on high-resource languages.
Since the 188 low-resource languages in WikiAnn are typologically and genealogically diverse, the improvement suggests that low-resource languages not only profit from cross-lingual transfer from similar languages \citep{cotterell2017crosslingual}, but that multilingual training brings other benefits, as well.
In multilingual training, certain aspects of the task at hand, such as tag distribution and BIO constraints have to be learned only once, while they have to be separately learned on each language in monolingual training.
Furthermore, multilingual training may prevent overfitting to biases in small monolingual datasets, such as a skewed tag distributions.
A visualization of the multilingual subword embedding space (Figure~\ref{fig:multi-emb-training}) gives evidence for this view.
Before training, distinct clusters of subword embeddings from the same language are visible.
After training, some of these clusters are more spread out and show more overlap, which indicates that some embeddings from different languages appear to have moved ``closer together'', as one would expect embeddings of semantically-related words to do.
However, the overall structure of the embedding space remains largely unchanged.
The model maintains language-specific subspaces and does not appear to create an interlingual semantic space which could facilitate cross-lingual transfer.

Having trained a multilingual model on all languages, we can further train this model on a single language (Table~\ref{tbl:wikiann-results-all}, \emph{+finetune}).
This finetuning further improves performance, giving the best overall score (91.4) and an 8.8 point improvement over \citeauthor{pan2017crosslingual} on low-resource languages (90.4 vs.\ 81.6).
These results show that \textbf{multilingual training followed by monolingual finetuning} is an effective method for low-resource sequence tagging.

\subsection{NER with Multilingual BERT}
\label{sec:ner-bert}

Table~\ref{tbl:wikiann-results-bert} shows NER results on the intersection of languages supported by all methods in our evaluation.
As in \secref{sec:ner-all}, FastText performs worst overall, monolingual BPEmb with character embeddings performs best on high-resource languages (93.6 F1), and multilingual BPEmb best on low-resource languages (91.1).
Multilingual BERT outperforms the \emph{Pan17} baseline and shows strong results in comparison to monolingual BPEmb. 
The combination of multilingual BERT, monolingual BPEmb, and character embeddings is best overall (92.0) among models trained only on monolingual NER data.
However, this ensemble of contextual and non-contextual subword embeddings is inferior to MultiBPEmb (93.2), which was first trained on multilingual data from all languages collectively, and then separately finetuned to each language.
Score distributions and detailed NER results for each language and method are shown in Appendix~\ref{sec:score-distributions} and Appendix~\ref{sec:detailer-ner-results}.

\label{sec:pos-tagging}
\begin{table*}
	\centering
	\small
	\begin{adjustbox}{max width=\linewidth}
		\input{tables_raw/ud_1_2_overall_results_highres_langs}
	\end{adjustbox}
	\caption{POS tagging accuracy on high-resource languages in UD 1.2.}
	\label{tbl:ud-results-all}
\end{table*}

\begin{table}
	\centering
	\small
	\begin{adjustbox}{max width=\linewidth}
		\input{tables_raw/ud_1_2_overall_results_lowres_langs}
	\end{adjustbox}
	\caption{POS tagging accuracy on low-resource languages in UD 1.2.}
	\label{tbl:ud-results-lowres}
\end{table}

\subsection{POS Tagging in 27 Languages}
We perform POS tagging experiments in the 21 high-resource (Table~\ref{tbl:ud-results-all}) and 6 low-resource languages (Table~\ref{tbl:ud-results-lowres}) from the Universal Dependencies (UD) treebanks on which \citet{yasunaga2018robust} report state-of-the-art results via adversarial training (\emph{Adv.}).
In high-resource POS tagging, we also compare to the \emph{BiLSTM} by \citet{plank2016multilingual}.
While differences between methods are less pronounced than for NER, we observe similar patterns.
On average, the combination of multilingual BERT, monolingual BPEmb, and character embeddings is best for high-resource languages and outperforms \emph{Adv.} by 0.2 percent (96.8 vs.\ 96.6).
For low-resource languages, multilingual BPEmb with character embeddings and finetuning is the best method, yielding an average improvement of 0.8 percent over \emph{Adv.} (92.4 vs.\ 91.6).

\section{Limitations and Conclusions}

\noindent\textbf{Limitations.} While extensive, our evaluation is not without limitations.
Throughout this study, we have used a Wikipedia edition in a given language as a sample of that language.
The degree to which this sample is representative varies, and low-resource Wikipedias in particular contain large fractions of ``foreign'' text and noise, which propagates into embeddings and datasets.
	Our evaluation did not include other subword representations, most notably ELMo \citep{peters2018contextualized} and contextual string embeddings \citep{akbik2018contextual}, since, even though they are language-agnostic in principle, pretrained models are only available in a few languages.

\noindent\textbf{Conclusions.} We have presented a large-scale study of contextual and non-contextual subword embeddings, in which we trained monolingual and multilingual NER models in 265 languages and POS-tagging models in 27 languages.
BPE vocabulary size has a large effect on model quality, both in monolingual settings and with a large vocabulary shared among 265 languages.
As a rule of thumb, a smaller vocabulary size is better for small datasets and larger vocabulary sizes better for larger datasets.
Large improvements over monolingual training showed that low-resource languages benefit from multilingual model training with shared subword embeddings.
Such improvements are likely not solely caused by cross-lingual transfer, but also by the prevention of overfitting and mitigation of noise in small monolingual datasets.
Monolingual finetuning of a multilingual model improves performance in almost all cases (compare \emph{-finetune} and \emph{+finetune} columns in Table~\ref{tbl:wikiann-full-results} in Appendix~\ref{sec:detailer-ner-results}).
For high-resource languages, we found that monolingual embeddings and monolingual training perform better than multilingual approaches with a shared vocabulary.
This is likely due to the fact that a high-resource language provides large background corpora for learning good embeddings of a large vocabulary and also provides so much training data for the task at hand that little additional information can be gained from training data in other languages.
Our experiments also show that even a large multilingual contextual model like BERT benefits from character embeddings and additional monolingual embeddings.

Finally, and while asking the reader to bear above limitations in mind, we make the following practical recommendations for multilingual sequence tagging with subword representations:
\begin{itemize}
	\item Choose the largest feasible subword vocabulary size when a large amount of data is available.
	\item Choose smaller subword vocabulary sizes in low-resource settings.
	\item Multilingual BERT is a robust choice across tasks and languages if the computational requirements can be met.
	\item With limited computational resources, use small monolingual, non-contextual representations, such as BPEmb combined with character embeddings.
	\item Combine different subword representations for better results.
	\item In low-resource scenarios, first perform multilingual pretraining with a shared subword vocabulary, then finetune to the language of interest.
\end{itemize}

\section{Acknowledgements}
We thank the anonymous reviewers for insightful comments.
This work has been funded by the Klaus Tschira Foundation, Heidelberg, Germany,
and partially funded by the German Research Foundation as part of the Research Training Group ``Adaptive Preparation of Information from Heterogeneous Sources'' (AIPHES) under grant No. GRK 1994/1. 

\bibliography{lit,add,arxiv}
\bibliographystyle{acl_natbib}

\newpage
\onecolumn
\twocolumn

\appendix

\newpage
\onecolumn
\section{Analysis of NER tag distribution and baseline performance in WikiAnn}
\label{sec:analysis-wikiann}
\begin{figure}[h!]
	\centering
	\includegraphics[width=\linewidth]{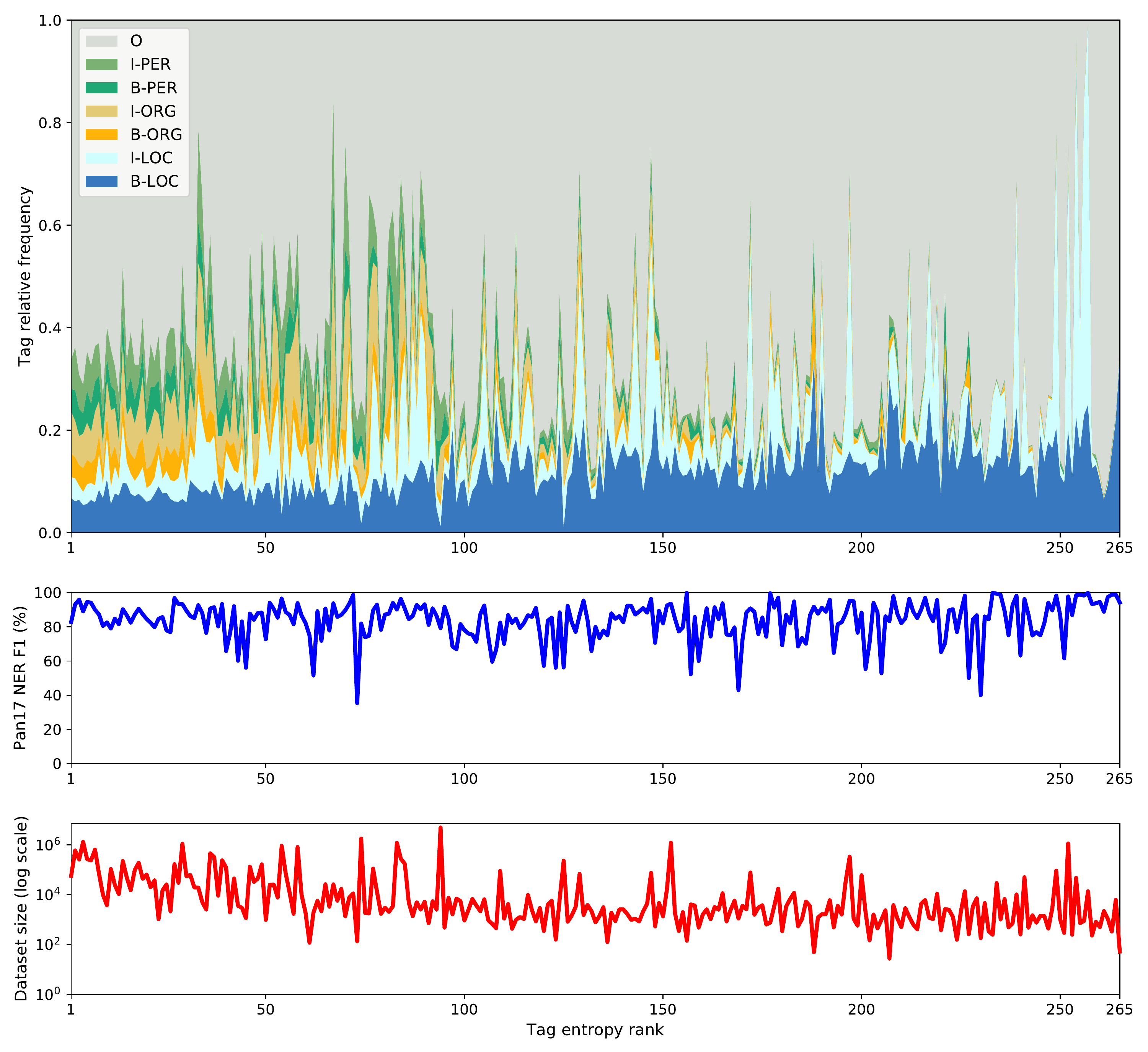}
	\caption{%
WikiAnn named entity tag distribution for each language (top) in comparison to \citeauthor{pan2017crosslingual} NER F1 scores (middle) and each language's dataset size (bottom).
Languages are sorted from left to right from highest to lowest tag distribution entropy.
That is, the NER tags in WikiAnn for the language in question are well-balanced for higher-ranked languages on the left and become more skewed for lower-ranked languages towards the right.
\citeauthor{pan2017crosslingual} achieve NER F1 scores up to 100 percent on some languages, which can be explained by the highly skewed, i.e.\,low-entropy, tag distribution in these languages (compare F1 scores \textgreater99\% in middle subfigure with skewed tag distributions in top subfigure). Better balance, i.e.\,higher entropy, of tag distribution tends to be found in languages for which WikiAnn provides more data (compare top and bottom subfigures).}
	\label{fig:tag-distributions}
\end{figure}

\newpage
\onecolumn
\twocolumn

\section{BPE and character-ngrams are not language-independent}
\label{app:language-independence}

Some methods proposed in NLP are unjustifiedly claimed to be language-independent \citep{bender2011independence}.
Subword segmentation with BPE or character-ngrams is language-agnostic, i.e., such a segmentation can be applied to any sequence of symbols, regardless of the language or meaning of these symbols.
However, BPE and character-ngrams are based on the assumption that meaningful subwords consist of adjacent characters, such as the suffix \emph{-ed} indicating past tense in English or the copular negation \emph{nai} in Japanese.
This assumption does not hold in languages with nonconcatenative morphology.
For example, Semitic roots in languages such as Arabic and Hebrew are patterns of discontinuous sequences of consonants which form words by insertion of vowels and other consonants. 
For instance, words related to \emph{writing} are derived from the root \emph{k-t-b}: \emph{kataba} ``he wrote'' or \emph{kitab} ``book''.
BPE and character-ngrams are not suited to efficiently capture such patterns of non-adjacent characters, and hence are not language-independent.

\section{Procedure for selecting the best BPE vocabulary size}
\label{sec:best-vocab-size}

We determine the best BPE vocabulary size for each language according to the following procedure.
\begin{enumerate}
	\item For each language $l$ in the set of all languages $L$ and each BPE vocabulary size $v \in V$, run $n$-fold cross-validation with each fold comprising a random split into training, development, and test set.\footnote{%
		$V = \{1000, 3000, 5000, 10000, 25000, 50000, 100000\}$ in our experiments.}
	\item Find the best BPE vocabulary size $v_l$ for each language, according to the mean evaluation score on the development set of each cross-validation fold.
	\item Determine the dataset size, measured in number of instances $N_l$, for each language.
	\item For each vocabulary size $v$, compute the median number of training instances of the languages for which $v$ gives the maximum evaluation score on the development set, i.e. $\widetilde{N}_v = \median(\{N_l | v = v_l \forall l \in L\})$.
	\item Given a language with dataset size $N_l$, the best BPE vocabulary size $\hat{v}_l$ is the one whose $\widetilde{N}_v$ is closest to $N_l$: $$\hat{v}_l = \argmin_{v \in V} \left| N_l - \widetilde{N}_v \right|$$
\end{enumerate}

\newpage
\onecolumn
\section{Sequence Tagging Model Hyper-Parameters}
\label{sec:hyper-parameters}
\begin{table}[h!]
	\centering
	\small
	\begin{tabular}{l|l|cc}
		\toprule
		 & & \multicolumn{2}{c}{Task} \\
		Subword method & Hyper-parameter & NER & POS \\
		\midrule
		FastText & Embedding dim. & 300 & 300 \\
		 & Encoder & biLSTM & biLSTM \\
		 & Encoder layer size & 256 & 256 \\
		 & Encoder layers & 2 & 2 \\
		 & Dropout & 0.5 & 0.2 \\
		 & Meta-LSTM layer size & 256 & 256 \\
		 & Meta-LSTM layers & 2 & 2 \\
		\midrule
		BPEmb & Embedding dim. & 100 & 100 \\
		 & Encoder & biLSTM & biLSTM \\
		 & Encoder layer size & 256 & 256 \\
		 & Encoder layers & 2 & 2 \\
		 & Dropout & 0.5 & 0.2 \\
		 & Char. embedding dim. & 50 & 50 \\
		 & Char. RNN layer size & 256 & 256 \\
		 & Shape embedding dim. & 50 & 50 \\
		 & Shape RNN layer size & 256 & 256 \\
		 & Meta-LSTM layer size & 256 & 256 \\
		 & Meta-LSTM layers & 2 & 2 \\
		\midrule
		MultiBPEmb & Embedding dim. & 300 & 300 \\
		 & Encoder & biLSTM & biLSTM \\
		 & Encoder layer size & 1024 & 1024 \\
		 & Encoder layers & 2 & 2 \\
		 & Dropout & 0.4 & 0.2 \\
		 & Char. embedding dim. & 100 & 100 \\
		 & Char. RNN layer size & 512 & 512 \\
		 & Meta-LSTM layer size & 1024 & 1024 \\
		 & Meta-LSTM layers & 2 & 2 \\
		\midrule
		BERT & Embedding dim. & 768 & 768 \\
		 & Encoder & Transformer & Transformer \\
		 & Encoder layer size & 768 & 768 \\
		 & Encoder layers & 12 & 12 \\
		 & Dropout & 0.2 & 0.2\\
		 & Char. embedding dim. & 50 & 50 \\
		 & Char. RNN layer size & 256 & 256 \\
		 & Meta-LSTM layer size & 256 & 256 \\
		 & Meta-LSTM layers & 2 & 2 \\
		\bottomrule
	\end{tabular}
	\caption{Hyper-parameters used in our experiments.}
	\label{tbl:hyper-params}
\end{table}

\onecolumn
\section{NER score distributions on WikiAnn}
\label{sec:score-distributions}
\begin{figure}[h!]
	\centering
	\includegraphics[width=0.87\linewidth]{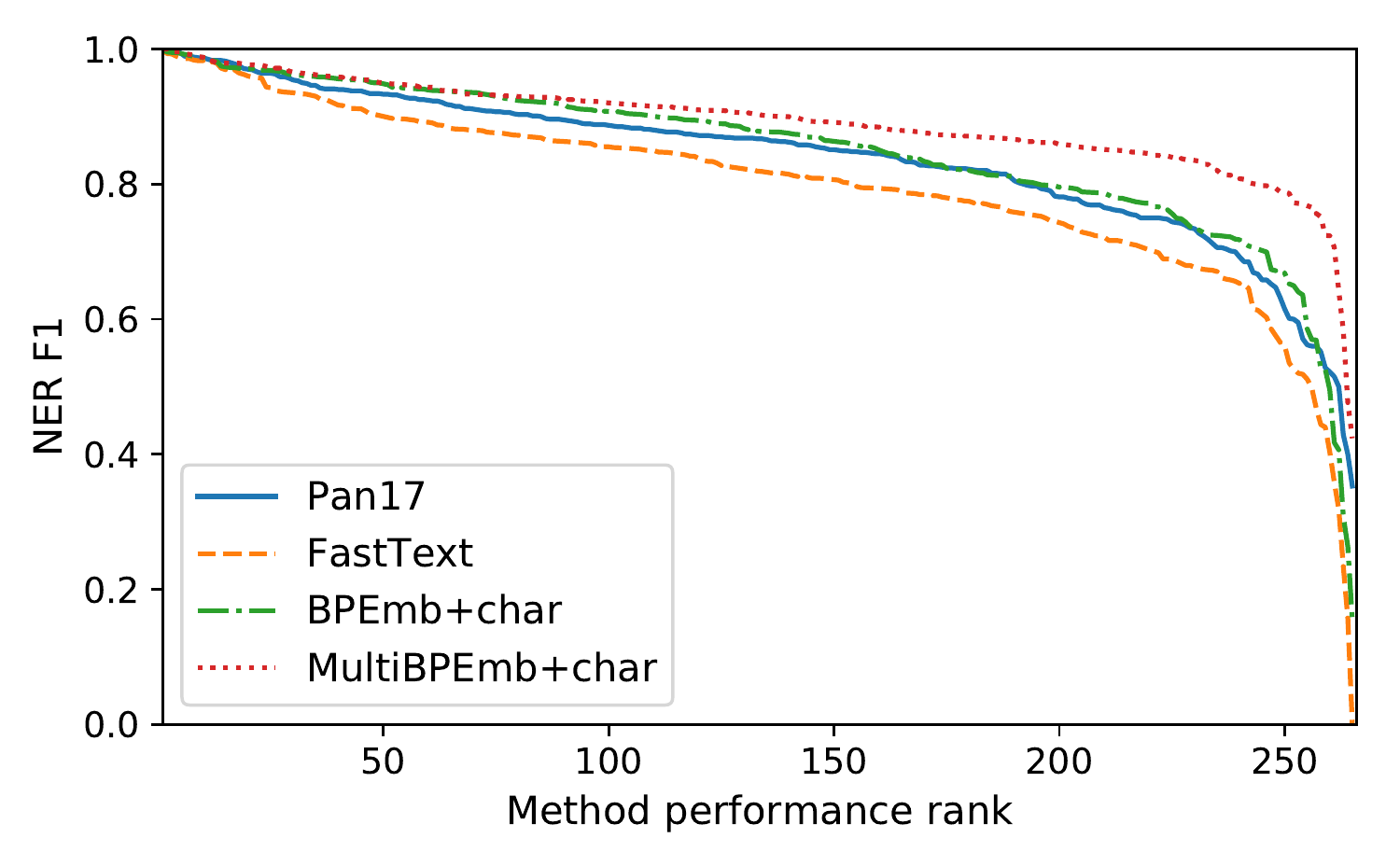}%

	\includegraphics[width=0.87\linewidth]{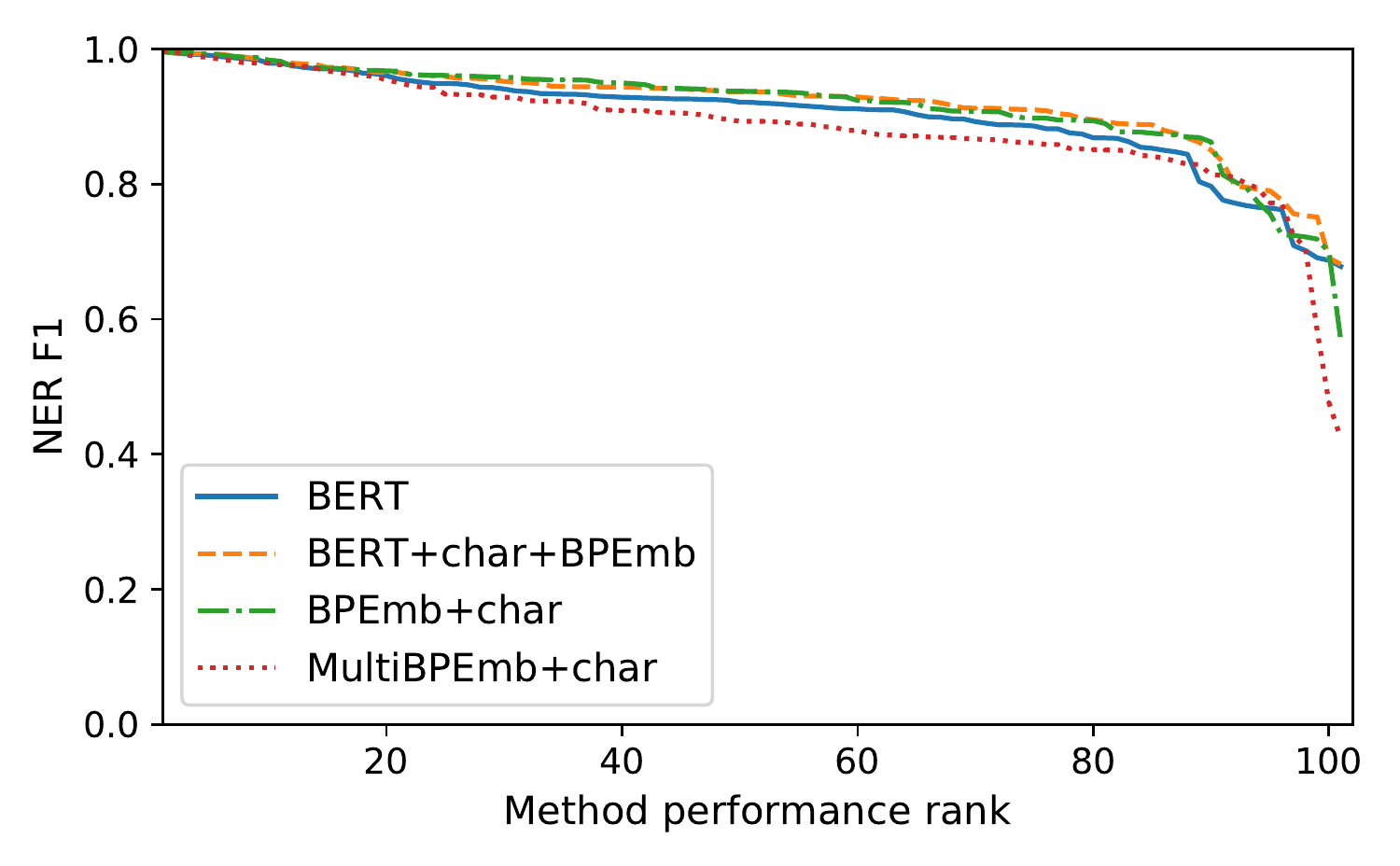}
	\caption{%
		NER results for the 265 languages represented in \citet{pan2017crosslingual}, FastText, and BPEmb (top), and the 101 languages constituting the intersection of these methods and BERT (bottom).
		Per-language F1 scores achieved by each method are sorted in descending order from left to right.
		The data points at rank 1 show the highest score among all languages achieved by the method in question, rank 2 the second-highest score etc.}
	\label{fig:sorted-lang-scores-all}
\end{figure}

\onecolumn
\newpage
\section{Detailed NER Results on WikiAnn}
\label{sec:detailer-ner-results}
\tiny
\centering
\input{tables_raw/wikiann_full_results}
\label{tbl:wikiann-full-results}

\end{document}

%% file: tables_raw/wikiann_overall_results_all_langs.tex
\begin{tabular}{lrr|rrrr|rr}
\toprule

 & & & \multicolumn{4}{|c|}{BPEmb} & \multicolumn{2}{|c}{MultiBPEmb+char} \\
Languages & Pan17 & FastText & BPEmb & +char & +shape & +someshape & -finetune & +finetune \\

\midrule
All (265) & 83.9 & 79.8 & 83.7 & 85.0 & 85.0 & 85.3 & 89.2 & \textbf{91.4}\\
Low-res. (188) & 81.6 & 76.7 & 79.7 & 81.4 & 81.5 & 81.9 & 89.7 & \textbf{90.4}\\
Med-res. (48) & 90.0 & 88.3 & 93.6 & 94.1 & 93.9 & 93.9 & 91.1 & \textbf{94.9}\\
High-res. (29) & 89.2 & 85.6 & 93.0 & \textbf{93.6} & 93.2 & 93.2 & 82.3 & 92.2\\
\bottomrule
\end{tabular}

%% file: tables_raw/wikiann_overall_results_bert_langs.tex
\begin{tabular}{lrr|rr|rrr}
\toprule

 & & & BPEmb & MultiBPEmb & \multicolumn{3}{|c}{BERT} \\
Languages & Pan17 & FastText & +char & +char+finetune & BERT & +char & +char+BPEmb \\

\midrule
All $\cap$ BERT (101) & 88.1 & 85.6 & 91.6 & \textbf{93.2} & 90.3 & 90.9 & 92.0\\
Low-res. $\cap$ BERT (27) & 83.6 & 81.3 & 85.1 & \textbf{91.1} & 85.4 & 85.6 & 87.1\\
Med-res. $\cap$ BERT (45) & 90.1 & 88.2 & 94.2 & \textbf{95.1} & 93.1 & 93.7 & 94.6\\
High-res. $\cap$ BERT (29) & 89.2 & 85.6 & \textbf{93.6} & 92.2 & 90.4 & 91.4 & 92.4\\
\bottomrule
\end{tabular}

%% file: tables_raw/ud_1_2_overall_results_highres_langs.tex
\begin{tabular}{lrrr|rrr|rrr|rr}
\toprule

 & & & & \multicolumn{3}{|c|}{BPEmb} & \multicolumn{3}{|c}{BERT} & \multicolumn{2}{|c}{MultiBPEmb+char}\\
Lang. & BiLSTM & Adv. & FastText & BPEmb & +char & +shape & BERT & +char & +char+BPemb & -finetune & +finetune \\

\midrule
Avg. & 96.4 & 96.6 & 95.6 & 95.2 & 96.4 & 95.7 & 95.6 & 96.3 & \textbf{96.8} & 96.1 & 96.6\\
\midrule
bg & 98.0 & 98.5 & 97.7 & 97.8 & 98.5 & 97.9 & 98.0 & 98.5 & \textbf{98.7} & 98.6 & \textbf{98.7}\\
cs & 98.2 & 98.8 & 98.3 & 98.5 & 98.9 & 98.7 & 98.4 & 98.8 & \textbf{99.0} & 97.9 & 98.9\\
da & 96.4 & 96.7 & 95.3 & 94.9 & 96.4 & 95.9 & 95.8 & 96.3 & \textbf{97.2} & 94.4 & 97.0\\
de & 93.4 & \textbf{94.4} & 90.8 & 92.7 & 93.8 & 93.5 & 93.7 & 93.8 & \textbf{94.4} & 93.6 & 94.0\\
en & 95.2 & 95.8 & 94.3 & 94.2 & 95.5 & 94.9 & 95.0 & 95.5 & \textbf{96.1} & 95.2 & 95.6\\
es & 95.7 & 96.4 & 96.3 & 96.1 & 96.6 & 96.0 & 96.1 & 96.3 & \textbf{96.8} & 96.4 & 96.5\\
eu & 95.5 & 94.7 & 94.6 & 94.3 & \textbf{96.1} & 94.8 & 93.4 & 95.0 & 96.0 & 95.3 & 95.6\\
fa & \textbf{97.5} & \textbf{97.5} & 97.1 & 95.9 & 97.0 & 96.0 & 95.7 & 96.5 & 97.3 & 97.0 & 97.1\\
fi & \textbf{95.8} & 95.4 & 92.8 & 92.8 & 94.4 & 93.5 & 92.1 & 93.8 & 94.3 & 92.2 & 94.6\\
fr & 96.1 & \textbf{96.6} & 96.0 & 95.5 & 96.1 & 95.8 & 96.1 & 96.5 & 96.5 & 96.2 & 96.2\\
he & 97.0 & \textbf{97.4} & 97.0 & 96.3 & 96.8 & 96.0 & 96.5 & 96.8 & 97.3 & 96.5 & 96.6\\
hi & 97.1 & 97.2 & 97.1 & 96.9 & 97.2 & 96.9 & 96.3 & 96.8 & \textbf{97.4} & 97.0 & 97.0\\
hr & \textbf{96.8} & 96.3 & 95.5 & 93.6 & 95.4 & 94.5 & 96.2 & 96.6 & \textbf{96.8} & 96.4 & \textbf{96.8}\\
id & 93.4 & \textbf{94.0} & 91.9 & 90.7 & 93.4 & 93.0 & 92.2 & 93.0 & 93.5 & 93.0 & 93.4\\
it & 98.0 & \textbf{98.1} & 97.4 & 97.0 & 97.8 & 97.3 & 97.5 & 97.9 & 98.0 & 97.9 & \textbf{98.1}\\
nl & 93.3 & 93.1 & 90.0 & 91.7 & 93.2 & 92.5 & 91.5 & 92.6 & 93.3 & 93.3 & \textbf{93.8}\\
no & 98.0 & 98.1 & 97.4 & 97.0 & 98.2 & 97.8 & 97.5 & 98.0 & \textbf{98.5} & 97.7 & 98.1\\
pl & 97.6 & 97.6 & 96.2 & 95.8 & 97.1 & 96.1 & 96.5 & \textbf{97.7} & 97.6 & 97.2 & 97.5\\
pt & 97.9 & 98.1 & 97.3 & 96.3 & 97.7 & 97.2 & 97.5 & 97.8 & 98.1 & 97.9 & \textbf{98.2}\\
sl & 96.8 & \textbf{98.1} & 97.1 & 96.2 & 97.7 & 96.8 & 96.3 & 97.4 & 97.9 & 97.7 & 98.0\\
sv & 96.7 & 96.7 & 96.7 & 95.3 & 96.7 & 95.7 & 96.2 & 97.1 & \textbf{97.4} & 96.7 & 97.3\\
\bottomrule
\end{tabular}

%% file: tables_raw/ud_1_2_overall_results_lowres_langs.tex
\begin{tabular}{lrr|r|r}
\toprule

 & & & BPEmb & MultiBPEmb \\
Lang. & Adv. & FastText & +char & +char+finetune \\

\midrule
Avg. & 91.6 & 90.4 & 79.3 & \textbf{92.4}\\
\midrule
el & \textbf{98.2} & 97.2 & 96.5 & 97.9\\
et & 91.3 & 89.5 & 82.1 & \textbf{92.8}\\
ga & \textbf{91.1} & 89.2 & 81.6 & 91.0\\
hu & \textbf{94.0} & 92.9 & 83.1 & \textbf{94.0}\\
ro & \textbf{91.5} & 88.6 & 73.9 & 89.7\\
ta & 83.2 & 85.2 & 58.7 & \textbf{88.7}\\
\bottomrule
\end{tabular}

%% file: tables_raw/wikiann_full_results.tex
\tablehead{
\toprule

 & & & & \multicolumn{3}{|c|}{BPEmb} & \multicolumn{3}{|c|}{BERT} & \multicolumn{2}{|c}{MultiBPEmb+char} \\
Language & \#inst. & Pan17 & FastText & BPEmb & +char & +shape & BERT & +char & +char+BPEmb & -finetune & +finetune\\

\midrule
}
\tabletail{\bottomrule}
\bottomcaption{Per-language NER F1 scores on WikiAnn.}
\begin{supertabular}{lrrr|rrr|rrr|rr}
ab & 474 & 60.0 & 76.3 & 69.2 & 83.9 & 77.8 & - & - & - & \textbf{85.4} & 83.3\\
ace & 3573 & 81.6 & 88.2 & 87.0 & 89.8 & 89.2 & - & - & - & \textbf{93.0} & \textbf{93.0}\\
ady & 693 & 92.7 & 82.2 & 86.3 & 90.9 & 91.9 & - & - & - & \textbf{96.3} & \textbf{96.3}\\
af & 14799 & 85.7 & 80.6 & 90.4 & 90.8 & 90.4 & 88.2 & 89.4 & 91.0 & 89.2 & \textbf{92.1}\\
ak & 244 & 86.8 & 68.9 & 72.5 & 89.5 & 75.8 & - & - & - & 91.3 & \textbf{94.1}\\
als & 7467 & 85.0 & 79.2 & 88.3 & 89.9 & 89.9 & - & - & - & 90.0 & \textbf{92.0}\\
am & 1032 & \textbf{84.7} & 35.8 & 62.1 & 66.8 & 67.2 & - & - & - & 75.7 & 76.3\\
an & 12719 & 93.0 & 82.7 & 94.1 & 93.9 & 94.7 & 95.1 & 95.9 & 96.6 & 94.4 & \textbf{97.0}\\
ang & 3848 & 84.0 & 75.2 & 79.8 & 78.4 & 80.4 & - & - & - & \textbf{84.8} & 84.7\\
ar & 164180 & 88.3 & 93.4 & 93.1 & \textbf{93.7} & 93.1 & 88.7 & 91.0 & 93.0 & 79.4 & 93.2\\
arc & 1618 & 68.5 & 65.8 & 78.7 & 79.5 & 76.2 & - & - & - & 84.1 & \textbf{85.6}\\
arz & 3256 & 77.8 & 81.7 & 78.0 & 78.8 & 76.5 & - & - & - & \textbf{85.7} & \textbf{85.7}\\
as & 1338 & 89.6 & \textbf{93.5} & 87.5 & 87.3 & 86.1 & - & - & - & 90.7 & 90.9\\
ast & 5598 & 89.2 & 82.1 & 89.8 & 89.5 & 90.3 & 91.2 & 92.1 & 92.4 & 94.6 & \textbf{94.9}\\
av & 1330 & 82.0 & 72.9 & 78.2 & 77.6 & 78.2 & - & - & - & 85.5 & \textbf{85.6}\\
ay & 7156 & 88.5 & 86.5 & 97.3 & 97.1 & 95.7 & - & - & - & \textbf{97.8} & 97.6\\
az & 19451 & 85.1 & 77.5 & 89.7 & 89.5 & 88.7 & 88.8 & 89.5 & 90.3 & 85.0 & \textbf{90.8}\\
azb & 2567 & 88.4 & 92.3 & 87.5 & 89.0 & 88.1 & 90.0 & 89.2 & 88.8 & 93.2 & \textbf{93.9}\\
ba & 11383 & 93.8 & 93.4 & 95.6 & 96.2 & 95.9 & 96.0 & 95.8 & 96.5 & 96.5 & \textbf{97.2}\\
bar & 17298 & 97.1 & 93.7 & 97.1 & 97.4 & 97.6 & 97.1 & 97.7 & 97.7 & 97.9 & \textbf{98.3}\\
bcl & 1047 & 82.3 & 75.4 & 74.0 & 74.4 & 74.1 & - & - & - & 91.2 & \textbf{92.9}\\
be & 32163 & 84.1 & 84.3 & 90.7 & 91.9 & 91.5 & 89.2 & 91.0 & \textbf{92.0} & 86.9 & \textbf{92.0}\\
bg & 121526 & 65.8 & 89.4 & 95.5 & \textbf{95.8} & 95.7 & 93.4 & 94.2 & 95.7 & 89.8 & 95.5\\
bi & 441 & 88.5 & 84.5 & 73.8 & 79.9 & 81.6 & - & - & - & \textbf{93.9} & \textbf{93.9}\\
bjn & 482 & 64.7 & 69.8 & 67.9 & 72.3 & 69.3 & - & - & - & 83.6 & \textbf{84.0}\\
bm & 345 & 77.3 & 67.1 & 63.3 & 64.0 & 71.2 & - & - & - & 79.8 & \textbf{80.8}\\
bn & 25898 & 93.8 & 96.0 & 95.9 & 95.8 & 95.9 & 95.3 & 95.2 & \textbf{96.6} & 92.2 & 96.3\\
bo & 2620 & 70.4 & 85.0 & \textbf{87.2} & 87.0 & 83.6 & - & - & - & 85.8 & 86.2\\
bpy & 876 & \textbf{98.3} & 96.4 & 95.2 & 96.8 & 95.6 & 97.0 & 95.2 & 94.4 & 97.9 & 97.9\\
br & 17003 & 87.0 & 82.2 & 90.6 & 92.1 & 91.1 & 89.7 & 90.6 & 92.7 & 89.6 & \textbf{93.1}\\
bs & 24191 & 84.8 & 80.6 & 88.1 & 89.8 & 89.2 & 89.6 & 89.8 & 90.9 & 88.0 & \textbf{92.1}\\
bug & 13676 & 99.9 & \textbf{100.0} & \textbf{100.0} & \textbf{100.0} & 99.9 & - & - & - & \textbf{100.0} & \textbf{100.0}\\
bxr & 2389 & 75.0 & 73.7 & 76.6 & 78.0 & 79.8 & - & - & - & 84.9 & \textbf{85.4}\\
ca & 222754 & 90.3 & 86.1 & 95.7 & \textbf{96.2} & 95.9 & 93.7 & 94.9 & 96.1 & 89.3 & 95.7\\
cdo & 2127 & \textbf{91.0} & 72.1 & 78.7 & 79.5 & 75.0 & - & - & - & 85.1 & 86.4\\
ce & 29027 & 99.4 & 99.3 & 99.5 & 99.6 & 99.5 & 99.7 & 99.7 & 99.7 & 99.6 & \textbf{99.8}\\
ceb & 50218 & 96.3 & 98.3 & 99.0 & 98.9 & 99.0 & 99.3 & 99.2 & 99.3 & 98.4 & \textbf{99.4}\\
ch & 146 & 70.6 & 40.3 & 39.7 & 67.4 & 60.0 & - & - & - & \textbf{78.8} & \textbf{78.8}\\
chr & 527 & 70.6 & 65.9 & 61.4 & 63.6 & 69.7 & - & - & - & 84.0 & \textbf{84.9}\\
chy & 405 & 85.1 & 77.6 & 77.3 & 81.1 & 75.8 & - & - & - & 86.2 & \textbf{88.5}\\
ckb & 5023 & 88.1 & 88.7 & 88.9 & 88.7 & 89.0 & - & - & - & 90.0 & \textbf{90.2}\\
co & 5654 & 85.4 & 74.5 & 86.4 & 83.9 & 84.7 & - & - & - & 91.6 & \textbf{92.3}\\
cr & 49 & \textbf{91.8} & 57.6 & 40.0 & 30.8 & 51.9 & - & - & - & 90.0 & 90.0\\
crh & 4308 & 90.1 & 88.2 & 90.6 & 92.6 & 91.3 & - & - & - & 93.0 & \textbf{93.3}\\
cs & 265794 & 94.6 & 85.7 & 94.3 & \textbf{95.0} & 94.7 & 92.7 & 93.8 & 94.3 & 85.0 & 94.5\\
csb & 3325 & 87.0 & 82.6 & 83.3 & 88.0 & 88.9 & - & - & - & 88.2 & \textbf{89.7}\\
cu & 842 & 75.5 & 68.0 & 74.4 & 81.8 & 78.0 & - & - & - & \textbf{87.0} & 85.6\\
cv & 10825 & 95.7 & 95.8 & 96.6 & 96.8 & 96.9 & \textbf{97.6} & 97.2 & 97.3 & 97.2 & 97.4\\
cy & 26039 & 90.7 & 86.1 & 92.9 & 93.8 & 93.6 & 91.6 & 92.8 & 93.0 & 90.5 & \textbf{94.4}\\
da & 95924 & 87.1 & 81.1 & 92.5 & 93.3 & 92.9 & 92.1 & 92.8 & \textbf{94.2} & 87.5 & 93.7\\
de & 1304068 & 89.0 & 77.2 & \textbf{94.4} & 93.0 & 94.1 & 88.8 & 89.6 & 91.2 & 80.1 & 90.6\\
diq & 1255 & 79.3 & 67.3 & 73.5 & 80.2 & 77.3 & - & - & - & 90.6 & \textbf{90.8}\\
dsb & 862 & 84.7 & 74.9 & 76.1 & 76.2 & 82.0 & - & - & - & 94.8 & \textbf{96.7}\\
dv & 1924 & 76.2 & 60.8 & 76.5 & 77.7 & 74.4 & - & - & - & 86.9 & \textbf{87.3}\\
dz & 258 & 50.0 & 51.8 & 88.2 & 80.5 & 76.2 & - & - & - & \textbf{93.3} & 91.4\\
ee & 252 & 63.2 & 64.5 & 54.4 & 56.9 & 57.8 & - & - & - & 87.8 & \textbf{90.5}\\
el & 63546 & 84.6 & 80.9 & 92.0 & 92.3 & 92.5 & 89.9 & 90.8 & \textbf{93.0} & 84.2 & 92.8\\
eo & 71700 & 88.7 & 84.7 & 93.7 & 94.3 & 94.2 & - & - & - & 88.1 & \textbf{94.8}\\
es & 811048 & 93.9 & 89.2 & 96.2 & \textbf{96.7} & 96.5 & 92.5 & 93.1 & 93.8 & 86.6 & 93.7\\
et & 48322 & 86.8 & 81.8 & 91.9 & 92.9 & 92.4 & 91.0 & 92.3 & \textbf{93.2} & 87.1 & \textbf{93.2}\\
eu & 89188 & 82.5 & 88.7 & 94.7 & 95.4 & 95.1 & 94.9 & 95.2 & \textbf{96.2} & 91.0 & 96.0\\
ext & 3141 & 77.8 & 71.6 & 78.3 & 78.8 & 78.8 & - & - & - & 85.4 & \textbf{87.4}\\
fa & 272266 & 96.4 & 97.2 & 96.9 & \textbf{97.3} & 96.8 & 94.7 & 95.3 & 96.1 & 86.7 & 96.2\\
ff & 154 & 76.9 & 52.0 & 68.2 & 72.4 & 76.7 & - & - & - & \textbf{90.9} & \textbf{90.9}\\
fi & 237372 & 93.4 & 81.5 & 93.1 & \textbf{93.7} & 93.2 & 91.2 & 92.0 & 93.1 & 82.9 & 92.8\\
fj & 125 & 75.0 & 49.8 & 65.9 & 52.7 & 52.4 & - & - & - & \textbf{100.0} & \textbf{100.0}\\
fo & 3968 & 83.6 & 82.4 & 85.1 & 87.7 & 87.1 & - & - & - & 92.0 & \textbf{92.2}\\
fr & 1095885 & 93.3 & 87.2 & 95.5 & \textbf{95.7} & 95.5 & 93.4 & 93.6 & 94.2 & 83.8 & 92.0\\
frp & 2358 & 86.2 & 86.9 & 86.6 & 89.6 & 90.4 & - & - & - & 93.4 & \textbf{94.7}\\
frr & 5266 & 70.1 & 79.5 & 86.7 & 88.2 & 88.6 & - & - & - & 90.1 & \textbf{91.1}\\
fur & 2487 & 84.5 & 77.1 & 79.7 & 78.6 & 81.4 & - & - & - & 86.3 & \textbf{88.3}\\
fy & 9822 & 86.6 & 80.7 & 89.8 & 90.8 & 90.5 & 88.2 & 89.3 & 90.4 & 91.9 & \textbf{93.0}\\
ga & 7569 & 85.3 & 77.6 & 87.3 & 87.8 & 86.8 & 85.5 & 86.4 & 86.2 & 89.1 & \textbf{92.0}\\
gag & 6716 & 89.3 & 91.2 & 94.9 & 96.9 & 95.3 & - & - & - & 96.2 & \textbf{97.5}\\
gan & 2876 & 84.9 & 79.6 & 87.3 & 88.1 & 85.8 & - & - & - & 91.9 & \textbf{92.0}\\
gd & 4906 & 92.8 & 81.6 & 85.5 & 86.4 & 87.7 & - & - & - & 92.4 & \textbf{93.5}\\
gl & 43043 & 87.4 & 78.7 & 92.8 & 93.7 & 93.1 & 92.7 & 93.2 & 93.9 & 90.2 & \textbf{94.9}\\
glk & 667 & 59.5 & \textbf{83.8} & 65.5 & 73.5 & 69.4 & - & - & - & 76.8 & 80.7\\
gn & 3689 & 71.2 & 72.3 & 82.1 & 79.9 & 81.1 & - & - & - & 83.5 & \textbf{85.4}\\
gom & 2192 & 88.8 & 93.6 & \textbf{95.8} & 95.6 & 95.4 & - & - & - & 92.7 & \textbf{95.8}\\
got & 475 & \textbf{91.7} & 61.3 & 62.8 & 70.2 & 67.8 & - & - & - & 81.4 & 82.6\\
gu & 2895 & 76.0 & 79.4 & 76.8 & 79.5 & 78.8 & 76.6 & 76.6 & \textbf{83.3} & 82.9 & 83.1\\
gv & 980 & 84.8 & 73.5 & 72.5 & 72.2 & 77.3 & - & - & - & 92.5 & \textbf{93.7}\\
ha & 489 & 75.0 & 85.5 & 82.9 & 82.8 & 81.3 & - & - & - & \textbf{94.7} & 93.8\\
hak & 3732 & 85.5 & 80.8 & 87.0 & 86.8 & 85.1 & - & - & - & 90.0 & \textbf{90.9}\\
haw & 1189 & 88.0 & 89.9 & 88.4 & 92.7 & 93.9 & - & - & - & 94.9 & \textbf{95.0}\\
he & 106569 & 79.0 & \textbf{91.6} & 90.8 & 91.2 & 90.6 & 84.8 & 88.4 & 91.3 & 70.6 & 88.9\\
hi & 11833 & 86.9 & 89.2 & 89.9 & 89.4 & 88.9 & 84.4 & 87.3 & 88.9 & 88.9 & \textbf{91.8}\\
hif & 715 & 81.1 & 76.8 & 71.6 & 77.2 & 78.7 & - & - & - & 95.6 & \textbf{96.1}\\
hr & 56235 & 82.8 & 80.9 & 89.5 & 90.7 & 90.5 & 90.3 & 90.6 & \textbf{92.4} & 86.5 & 91.8\\
hsb & 3181 & 91.5 & 91.7 & 88.3 & 90.4 & 91.7 & - & - & - & \textbf{95.9} & 95.8\\
ht & 6166 & 98.9 & 99.0 & 98.8 & 99.1 & 98.8 & 98.6 & 99.0 & 98.8 & 99.6 & \textbf{99.7}\\
hu & 253111 & \textbf{95.9} & 85.3 & 95.0 & 95.4 & 95.2 & 92.4 & 93.1 & 94.4 & 86.3 & 94.7\\
hy & 25106 & 90.4 & 85.0 & 93.2 & 93.6 & 93.5 & 92.0 & 92.7 & 93.7 & 89.3 & \textbf{94.4}\\
ia & 6672 & 75.4 & 79.3 & 81.3 & 84.2 & 84.7 & - & - & - & 88.5 & \textbf{89.9}\\
id & 131671 & 87.8 & 85.4 & 94.5 & 95.1 & 94.7 & 93.3 & 93.7 & 94.9 & 89.3 & \textbf{95.4}\\
ie & 1645 & 88.8 & 85.6 & 90.3 & 90.0 & 87.4 & - & - & - & 95.2 & \textbf{95.7}\\
ig & 937 & 74.4 & 68.9 & 82.7 & 83.4 & 83.6 & - & - & - & 88.9 & \textbf{89.5}\\
ik & 431 & \textbf{94.1} & 83.1 & 88.6 & 89.3 & 89.2 & - & - & - & 93.3 & 93.8\\
ilo & 2511 & 90.3 & 80.9 & 87.6 & 81.2 & 86.1 & - & - & - & 95.8 & \textbf{96.3}\\
io & 2979 & 87.2 & 86.4 & 88.1 & 87.4 & 90.8 & 91.1 & 92.0 & 92.5 & 95.4 & \textbf{95.8}\\
is & 8978 & 80.2 & 75.7 & 85.6 & 87.0 & 87.1 & 86.8 & 83.8 & 87.5 & 88.4 & \textbf{90.7}\\
it & 909085 & \textbf{96.6} & 89.6 & 96.1 & 96.1 & 96.3 & 93.8 & 93.7 & 94.5 & 87.1 & 94.0\\
iu & 447 & 66.7 & 68.6 & 84.0 & 88.9 & 86.6 & - & - & - & \textbf{92.8} & 92.3\\
ja & 4902623 & \textbf{79.2} & 71.0 & 67.7 & 71.9 & 68.9 & 67.8 & 69.0 & 69.1 & 47.6 & 68.4\\
jbo & 1669 & 92.4 & 87.9 & 89.0 & 90.6 & 88.7 & - & - & - & 94.4 & \textbf{94.5}\\
jv & 3719 & 82.6 & 67.4 & 83.6 & 87.3 & 87.1 & 87.6 & 88.1 & 89.0 & 92.3 & \textbf{93.2}\\
ka & 37500 & 79.8 & 89.0 & 89.5 & 89.4 & 88.5 & 85.3 & 87.6 & \textbf{89.7} & 81.4 & 89.3\\
kaa & 1929 & 55.2 & 77.2 & 78.4 & 81.3 & 82.0 & - & - & - & 88.5 & \textbf{89.4}\\
kab & 3004 & 75.7 & 79.4 & 85.8 & 86.1 & 86.5 & - & - & - & 87.9 & \textbf{89.1}\\
kbd & 1482 & 74.9 & 74.3 & 81.3 & 83.7 & 84.8 & - & - & - & 90.4 & \textbf{91.6}\\
kg & 1379 & 82.1 & 93.0 & 91.8 & 93.8 & \textbf{95.7} & - & - & - & 95.4 & 95.6\\
ki & 1056 & \textbf{97.5} & 93.6 & 91.9 & 93.5 & 93.3 & - & - & - & 97.2 & 97.2\\
kk & 60248 & 88.3 & 93.8 & 97.0 & 97.5 & 97.1 & 97.3 & 97.3 & \textbf{97.8} & 95.9 & 97.6\\
kl & 1403 & 75.0 & 86.4 & 83.6 & 85.9 & 88.8 & - & - & - & \textbf{92.9} & 92.6\\
km & 4036 & 52.2 & 51.1 & 87.1 & 85.6 & 85.6 & - & - & - & \textbf{91.2} & 90.7\\
kn & 3567 & 60.1 & 76.0 & 72.4 & 77.3 & 74.5 & 68.7 & 71.4 & 75.1 & \textbf{81.3} & 80.5\\
ko & 188823 & 90.6 & 44.4 & 91.5 & \textbf{92.1} & 91.7 & 86.8 & 88.4 & 91.1 & 72.4 & 90.6\\
koi & 2798 & 89.6 & 90.2 & 91.2 & 92.0 & 92.0 & - & - & - & 93.0 & \textbf{93.7}\\
krc & 1830 & 84.9 & 75.6 & 78.2 & 82.3 & 83.4 & - & - & - & \textbf{89.8} & 89.1\\
ks & 117 & \textbf{75.0} & 23.4 & 23.8 & 40.7 & 34.1 & - & - & - & 64.2 & 64.2\\
ksh & 1138 & 56.0 & 44.0 & 57.6 & 52.6 & 60.2 & - & - & - & 72.4 & \textbf{74.1}\\
ku & 2953 & 83.2 & 71.1 & 79.3 & 81.2 & 85.2 & - & - & - & 90.9 & \textbf{91.7}\\
kv & 2464 & 89.7 & 85.3 & 83.1 & 85.0 & 84.9 & - & - & - & 93.1 & \textbf{94.1}\\
kw & 1587 & 94.0 & 90.4 & 90.4 & 91.1 & 92.7 & - & - & - & 97.1 & \textbf{97.7}\\
ky & 2153 & 71.8 & 58.6 & 67.2 & 69.9 & 72.9 & 70.9 & 72.9 & 75.3 & 81.0 & \textbf{82.0}\\
la & 77279 & 90.8 & 93.1 & 96.2 & 97.1 & 97.0 & 96.8 & 97.1 & \textbf{97.3} & 92.8 & 97.1\\
lad & 973 & 92.3 & 79.5 & 80.0 & 82.8 & 83.0 & - & - & - & 93.9 & \textbf{94.1}\\
lb & 10450 & 81.5 & 68.0 & 87.3 & 86.9 & 86.6 & 86.3 & 86.4 & 88.8 & 86.2 & \textbf{89.7}\\
lbe & 631 & 88.9 & 81.1 & 84.4 & 84.5 & 86.2 & - & - & - & 91.8 & \textbf{92.6}\\
lez & 3310 & 84.2 & 87.6 & 89.2 & 90.4 & 91.2 & - & - & - & 93.8 & \textbf{94.2}\\
lg & 328 & \textbf{98.8} & 92.0 & 91.5 & 91.3 & 91.0 & - & - & - & 97.2 & 97.2\\
li & 4634 & 89.4 & 83.4 & 86.3 & 90.4 & 88.0 & - & - & - & 93.7 & \textbf{94.9}\\
lij & 3546 & 72.3 & 75.9 & 79.9 & 82.2 & 82.3 & - & - & - & 87.3 & \textbf{87.5}\\
lmo & 13715 & 98.3 & 98.6 & 98.5 & 98.8 & 99.0 & 99.1 & \textbf{99.3} & \textbf{99.3} & 98.8 & \textbf{99.3}\\
ln & 1437 & 82.8 & 68.3 & 74.3 & 81.3 & 78.8 & - & - & - & 87.2 & \textbf{87.4}\\
lo & 991 & 52.8 & 67.7 & 70.5 & 76.6 & 72.6 & - & - & - & 86.1 & \textbf{86.8}\\
lrc & 372 & 65.2 & 70.5 & 59.3 & 71.8 & 66.0 & - & - & - & 79.8 & \textbf{80.0}\\
lt & 60871 & 86.3 & 84.1 & 91.2 & 92.4 & 91.4 & 90.7 & 91.5 & \textbf{92.7} & 85.9 & 92.2\\
ltg & 1036 & 74.3 & 78.3 & 80.6 & 82.1 & 82.8 & - & - & - & 88.8 & \textbf{89.0}\\
lv & 44434 & 92.1 & 87.6 & 92.7 & 94.1 & 93.9 & 91.9 & 93.1 & \textbf{94.2} & 87.2 & 94.0\\
mai & 755 & 99.7 & 98.1 & 98.4 & 98.3 & 98.4 & - & - & - & 99.6 & \textbf{100.0}\\
mdf & 497 & 82.2 & 65.3 & 71.6 & 74.9 & 76.0 & - & - & - & 84.2 & \textbf{88.4}\\
mg & 11181 & 98.7 & 99.3 & 99.4 & 99.3 & 99.4 & 99.4 & 99.4 & 99.4 & 99.1 & \textbf{99.5}\\
mhr & 3443 & 86.7 & 88.4 & 89.0 & 92.2 & 89.9 & - & - & - & 94.8 & \textbf{95.3}\\
mi & 5980 & 95.9 & 92.6 & 96.2 & 96.5 & 96.1 & - & - & - & 96.4 & \textbf{97.6}\\
min & 3626 & 85.8 & 84.5 & 87.9 & 87.7 & 88.3 & 86.8 & 89.8 & 91.2 & 94.3 & \textbf{94.6}\\
mk & 29421 & 93.4 & 87.4 & 93.6 & 94.2 & 94.0 & 92.9 & 92.5 & 93.7 & 90.6 & \textbf{94.6}\\
ml & 19729 & 82.4 & \textbf{86.3} & 84.7 & 86.2 & 84.6 & 79.7 & 81.5 & 85.0 & 77.2 & 84.2\\
mn & 2511 & 76.4 & 71.2 & 73.1 & 72.5 & 77.6 & 76.8 & 76.0 & 79.5 & 85.9 & \textbf{87.0}\\
mr & 14978 & 82.4 & 88.0 & 86.8 & 87.7 & 87.1 & 85.0 & 85.9 & 88.0 & 85.0 & \textbf{89.7}\\
mrj & 6036 & 97.0 & 96.9 & 96.8 & 96.9 & 97.6 & - & - & - & 97.7 & \textbf{98.3}\\
ms & 67867 & 86.8 & 88.0 & 95.4 & 95.9 & 95.4 & 94.9 & 95.4 & 95.9 & 92.3 & \textbf{96.7}\\
mt & 1883 & 82.3 & 68.9 & 77.1 & 80.1 & 78.9 & - & - & - & 84.5 & \textbf{87.0}\\
mwl & 2410 & 76.1 & 65.1 & 75.4 & 73.7 & 73.4 & - & - & - & 80.0 & \textbf{80.8}\\
my & 1908 & 51.5 & 73.3 & 72.2 & 72.2 & 70.5 & 69.1 & 72.4 & 75.6 & \textbf{77.1} & 76.3\\
myv & 2108 & 88.6 & 90.3 & 86.7 & 90.3 & 90.0 & - & - & - & 92.9 & \textbf{93.2}\\
mzn & 2491 & 86.4 & 89.2 & 88.5 & 87.7 & 86.6 & - & - & - & 91.8 & \textbf{92.2}\\
na & 1107 & 87.6 & 84.7 & 83.7 & 88.6 & 90.0 & - & - & - & 94.4 & \textbf{95.2}\\
nap & 4205 & 86.9 & 72.4 & 81.5 & 82.1 & 80.7 & - & - & - & 87.7 & \textbf{88.7}\\
nds & 4798 & 84.5 & 78.0 & 87.4 & 90.1 & 89.3 & 88.6 & 88.9 & 89.5 & 93.2 & \textbf{93.3}\\
ne & 1685 & 81.5 & 80.2 & 79.3 & 75.6 & 74.2 & 76.2 & 77.1 & 79.7 & \textbf{87.9} & 87.7\\
new & 10163 & 98.2 & 98.6 & 98.3 & 98.2 & 98.3 & 97.9 & 98.4 & 98.3 & 98.8 & \textbf{99.5}\\
nl & 589714 & 93.2 & 85.2 & 94.4 & \textbf{95.5} & 95.3 & 92.6 & 92.5 & 93.5 & 86.9 & 93.5\\
nn & 44228 & 88.1 & 85.3 & 93.6 & 94.7 & 94.2 & 93.3 & 93.4 & 94.5 & 90.6 & \textbf{95.0}\\
no & 233037 & 94.1 & 86.9 & 94.8 & \textbf{95.4} & 95.0 & 93.2 & 93.6 & 95.0 & 87.0 & 94.8\\
nov & 3176 & 77.0 & 87.2 & 94.0 & 94.3 & 93.5 & - & - & - & 97.9 & \textbf{98.0}\\
nrm & 1281 & 96.4 & 89.7 & 88.1 & 91.9 & 92.4 & - & - & - & 97.9 & \textbf{98.3}\\
nso & 720 & 98.9 & 98.7 & 97.2 & 97.2 & 97.7 & - & - & - & \textbf{99.2} & 99.1\\
nv & 2569 & 90.9 & 81.7 & 80.2 & 83.2 & 83.0 & - & - & - & \textbf{91.6} & 90.7\\
ny & 156 & 56.0 & 46.8 & 48.0 & 41.7 & 40.8 & - & - & - & \textbf{86.1} & \textbf{86.1}\\
oc & 16915 & 92.5 & 87.7 & 93.0 & 93.1 & 94.6 & 94.3 & 94.4 & 95.2 & 93.3 & \textbf{96.5}\\
om & 631 & 74.2 & 67.2 & 69.9 & 72.8 & 75.6 & - & - & - & 78.8 & \textbf{80.6}\\
or & 1362 & 86.4 & 75.6 & 86.6 & 84.0 & 82.2 & - & - & - & 92.5 & \textbf{93.0}\\
os & 2155 & 87.4 & 81.2 & 82.4 & 85.5 & 84.7 & - & - & - & 91.4 & \textbf{91.6}\\
pa & 1773 & 74.8 & 81.9 & 75.2 & 72.4 & 77.7 & 77.6 & 74.8 & 79.0 & \textbf{85.3} & 84.8\\
pag & 1643 & 91.2 & 89.5 & 87.2 & 88.6 & 89.9 & - & - & - & \textbf{91.5} & 91.2\\
pam & 1072 & 87.2 & 78.4 & 76.8 & 78.0 & 84.3 & - & - & - & 93.1 & \textbf{93.5}\\
pap & 1555 & \textbf{88.8} & 72.7 & 79.0 & 76.4 & 80.7 & - & - & - & 87.5 & 87.1\\
pcd & 4591 & 86.1 & 86.9 & 88.1 & 91.4 & 90.3 & - & - & - & 91.4 & \textbf{92.2}\\
pdc & 1571 & 78.1 & 71.6 & 75.7 & 79.7 & 80.5 & - & - & - & 84.7 & \textbf{87.0}\\
pfl & 1092 & 42.9 & 56.6 & 62.3 & 65.0 & 64.9 & - & - & - & 76.5 & \textbf{78.9}\\
pi & 27 & 83.3 & 0.0 & 25.0 & 15.4 & 0.0 & - & - & - & \textbf{90.9} & \textbf{90.9}\\
pih & 470 & 87.2 & 78.5 & 73.1 & 76.7 & 86.0 & - & - & - & \textbf{91.8} & \textbf{91.8}\\
pl & 639987 & 90.0 & 86.0 & 94.4 & \textbf{95.0} & 94.5 & 91.0 & 91.4 & 92.9 & 84.2 & 92.6\\
pms & 3809 & 98.0 & 95.7 & 96.4 & 96.1 & 96.1 & 97.0 & 97.3 & 97.9 & 97.9 & \textbf{98.2}\\
pnb & 5471 & 90.8 & 91.2 & 90.2 & 89.8 & 90.7 & 91.4 & 90.1 & 91.2 & 90.9 & \textbf{91.7}\\
pnt & 291 & 61.5 & 70.1 & 66.2 & 71.3 & 73.5 & - & - & - & 77.2 & \textbf{78.3}\\
ps & 6888 & 66.9 & 79.2 & 77.8 & 77.9 & 77.4 & - & - & - & 78.6 & \textbf{79.8}\\
pt & 452130 & 90.7 & 86.3 & 95.7 & \textbf{96.0} & 95.8 & 92.6 & 92.8 & 93.7 & 86.8 & 94.3\\
qu & 6480 & 92.5 & 90.0 & 93.2 & 93.9 & 93.3 & - & - & - & 96.0 & \textbf{97.1}\\
rm & 6617 & 82.0 & 80.3 & 86.2 & 87.8 & 87.1 & - & - & - & 90.1 & \textbf{91.0}\\
rmy & 532 & 68.5 & 65.6 & 80.4 & 81.3 & 80.8 & - & - & - & \textbf{93.0} & \textbf{93.0}\\
rn & 179 & 40.0 & 52.6 & 65.7 & 65.2 & 82.6 & - & - & - & \textbf{94.7} & \textbf{94.7}\\
ro & 171314 & 90.6 & 87.6 & 95.7 & \textbf{96.8} & 95.6 & 94.8 & 94.7 & 95.6 & 90.4 & 96.4\\
ru & 1192873 & 90.1 & 89.7 & 95.2 & \textbf{95.4} & 94.7 & 91.8 & 92.0 & 93.0 & 85.1 & 92.2\\
rue & 1583 & 82.7 & 78.1 & 76.0 & 81.7 & 84.2 & - & - & - & 89.1 & \textbf{89.8}\\
rw & 1517 & \textbf{95.4} & 86.2 & 83.9 & 89.1 & 87.6 & - & - & - & 92.7 & 93.3\\
sa & 1827 & 73.9 & 76.7 & 78.4 & 78.7 & 71.4 & - & - & - & \textbf{80.8} & 80.6\\
sah & 3442 & 91.2 & 89.6 & 91.5 & 92.2 & 91.1 & - & - & - & \textbf{95.0} & 94.6\\
sc & 917 & 78.1 & 74.6 & 71.9 & 70.8 & 76.4 & - & - & - & \textbf{86.9} & 86.6\\
scn & 5181 & 93.2 & 82.6 & 88.9 & 91.1 & 90.7 & 91.5 & 91.6 & 92.4 & 95.0 & \textbf{95.2}\\
sco & 9714 & 86.8 & 84.1 & 88.9 & 90.7 & 90.7 & 89.0 & 89.8 & 91.1 & 90.8 & \textbf{93.2}\\
sd & 2186 & 65.8 & 80.1 & 78.7 & 81.7 & 75.2 & - & - & - & 82.0 & \textbf{84.9}\\
se & 1256 & 90.3 & 92.6 & 88.6 & 91.0 & 91.8 & - & - & - & 95.7 & \textbf{95.8}\\
sg & 245 & \textbf{99.9} & 71.5 & 92.0 & 86.2 & 93.2 & - & - & - & 96.0 & 96.0\\
sh & 1126257 & 97.8 & 98.1 & 99.4 & \textbf{99.5} & 99.4 & 98.8 & 98.9 & 98.9 & 98.3 & 99.1\\
si & 2025 & \textbf{87.7} & 87.0 & 80.2 & 80.3 & 79.4 & - & - & - & 85.2 & 87.3\\
sk & 68845 & 87.3 & 83.5 & 92.4 & 93.5 & 93.1 & 92.9 & 93.7 & 94.4 & 88.5 & \textbf{94.5}\\
sl & 54515 & 89.5 & 86.2 & 93.0 & 94.2 & 93.8 & 93.0 & 94.4 & 95.1 & 90.9 & \textbf{95.2}\\
sm & 773 & 80.0 & 56.0 & 65.5 & 70.4 & 64.2 & - & - & - & 80.7 & \textbf{81.9}\\
sn & 1064 & \textbf{95.0} & 71.6 & 79.7 & 79.3 & 80.7 & - & - & - & 89.3 & 89.7\\
so & 5644 & 85.8 & 75.3 & 82.6 & 84.5 & 84.5 & - & - & - & 88.0 & \textbf{89.3}\\
sq & 24602 & 94.1 & 85.5 & 93.2 & 94.2 & 94.2 & 94.3 & 94.8 & 95.5 & 93.3 & \textbf{95.7}\\
sr & 331973 & 95.3 & 94.3 & 96.8 & \textbf{97.1} & \textbf{97.1} & 96.4 & 96.3 & 96.8 & 92.9 & 96.6\\
srn & 568 & 76.5 & 81.9 & 89.4 & 90.3 & 88.2 & - & - & - & 93.8 & \textbf{94.6}\\
ss & 341 & 69.2 & 74.1 & 81.9 & 77.2 & 82.6 & - & - & - & 87.4 & \textbf{88.0}\\
st & 339 & 84.4 & 78.6 & 88.2 & 93.3 & 91.1 & - & - & - & \textbf{96.6} & \textbf{96.6}\\
stq & 1085 & 70.0 & 76.6 & 78.9 & 77.4 & 74.1 & - & - & - & 91.4 & \textbf{91.9}\\
su & 960 & 72.7 & 53.5 & 58.8 & 57.0 & 66.8 & 76.4 & 69.6 & 68.1 & 87.3 & \textbf{89.0}\\
sv & 1210937 & 93.6 & 96.2 & 98.5 & \textbf{98.8} & 98.7 & 97.9 & 98.0 & 98.1 & 96.8 & 97.8\\
sw & 7589 & 93.4 & 85.2 & 91.0 & 90.7 & 90.8 & 91.0 & 91.7 & 91.7 & 92.8 & \textbf{93.6}\\
szl & 2566 & 82.7 & 77.9 & 79.6 & 82.2 & 84.1 & - & - & - & 92.1 & \textbf{93.1}\\
ta & 25663 & 77.9 & \textbf{86.3} & 84.5 & 85.7 & 84.3 & - & - & - & 75.2 & 84.2\\
te & 9929 & 80.5 & \textbf{87.9} & 87.8 & 87.5 & 87.5 & 80.4 & 83.7 & 86.8 & 83.4 & 87.5\\
tet & 1051 & 73.5 & 79.3 & 81.1 & 85.3 & 84.0 & - & - & - & 92.8 & \textbf{93.0}\\
tg & 4277 & 88.3 & 85.4 & 89.6 & 89.8 & 88.8 & 87.4 & 88.4 & 89.3 & 92.3 & \textbf{94.1}\\
th & 230508 & 56.2 & 81.0 & 80.8 & 81.4 & \textbf{81.6} & 70.2 & 78.4 & 77.6 & 42.4 & 77.7\\
ti & 52 & \textbf{94.2} & 60.2 & 77.3 & 49.5 & 32.9 & - & - & - & 91.7 & 91.7\\
tk & 2530 & 86.3 & 81.5 & 82.7 & 82.8 & 83.7 & - & - & - & 89.0 & \textbf{89.8}\\
tl & 19109 & 92.7 & 79.4 & 93.9 & 93.7 & 93.7 & 92.8 & 94.2 & 94.0 & 92.2 & \textbf{96.2}\\
tn & 750 & 76.9 & 72.6 & 72.3 & 79.8 & 81.2 & - & - & - & 83.6 & \textbf{84.7}\\
to & 814 & \textbf{92.3} & 77.0 & 67.6 & 74.9 & 81.2 & - & - & - & 86.3 & 88.2\\
tpi & 1038 & 83.3 & 84.7 & 84.6 & 86.4 & 88.5 & - & - & - & 94.7 & \textbf{95.6}\\
tr & 167272 & \textbf{96.9} & 77.5 & 94.4 & 94.9 & 94.5 & 92.6 & 93.1 & 94.4 & 86.1 & 95.1\\
ts & 227 & 93.3 & \textbf{94.4} & 78.9 & 86.3 & 77.0 & - & - & - & 91.3 & 92.2\\
tt & 35174 & 87.7 & 96.9 & 98.4 & 98.4 & 98.4 & 98.4 & 98.2 & 98.6 & 97.7 & \textbf{98.8}\\
tum & 815 & 93.8 & 95.8 & 90.7 & 93.7 & 93.2 & - & - & - & \textbf{97.6} & \textbf{97.6}\\
tw & 491 & 94.6 & 91.2 & 87.5 & 92.3 & 94.8 & - & - & - & \textbf{97.9} & \textbf{97.9}\\
ty & 1004 & 86.7 & 90.8 & \textbf{97.2} & 94.3 & 96.0 & - & - & - & 95.4 & 95.6\\
tyv & 842 & \textbf{91.1} & 70.3 & 73.4 & 67.2 & 65.0 & - & - & - & 84.6 & 84.5\\
udm & 840 & 88.9 & 83.4 & 85.6 & 85.6 & 83.6 & - & - & - & 95.6 & \textbf{96.6}\\
ug & 1998 & 79.7 & 84.6 & 83.2 & 82.0 & 80.0 & - & - & - & 87.1 & \textbf{87.4}\\
uk & 319693 & 91.5 & 91.2 & 95.6 & \textbf{96.0} & 95.8 & 92.1 & 92.5 & 93.7 & 88.9 & 94.9\\
ur & 74841 & 96.4 & 96.9 & 97.0 & 97.1 & 97.0 & 95.6 & 96.6 & 97.1 & 91.0 & \textbf{97.3}\\
uz & 91284 & 98.3 & 97.9 & 99.0 & \textbf{99.3} & 99.2 & 99.2 & \textbf{99.3} & \textbf{99.3} & 97.6 & \textbf{99.3}\\
ve & 141 & \textbf{99.9} & 31.8 & 21.0 & 58.6 & 73.0 & - & - & - & 89.2 & 89.2\\
vec & 1861 & 87.9 & 78.3 & 80.3 & 84.8 & 82.7 & - & - & - & 92.9 & \textbf{93.0}\\
vep & 2406 & 85.8 & 87.1 & 88.8 & 89.0 & 89.3 & - & - & - & 92.0 & \textbf{93.2}\\
vi & 110535 & 89.6 & 88.1 & 93.4 & 94.1 & 93.8 & 92.5 & 93.4 & 94.4 & 85.2 & \textbf{94.8}\\
vls & 1683 & 78.2 & 70.7 & 78.2 & 78.7 & 78.7 & - & - & - & 83.8 & \textbf{84.5}\\
vo & 46876 & 98.5 & 98.3 & 99.1 & 99.5 & 99.3 & 98.7 & 99.1 & 99.2 & 97.4 & \textbf{99.7}\\
wa & 5503 & 81.6 & 78.9 & 84.6 & 83.7 & 84.4 & - & - & - & \textbf{87.1} & 87.0\\
war & 11748 & 94.9 & 93.3 & 95.4 & 95.5 & 95.9 & 96.3 & 96.1 & 95.7 & 96.1 & \textbf{97.8}\\
wo & 1196 & \textbf{87.7} & 82.3 & 79.1 & 79.4 & 78.5 & - & - & - & 84.6 & 86.5\\
wuu & 5683 & 79.7 & 67.5 & 87.0 & 87.6 & 86.7 & - & - & - & 91.5 & \textbf{92.5}\\
xal & 1005 & 98.7 & 98.4 & 95.8 & 95.6 & 95.9 & - & - & - & \textbf{99.3} & 98.9\\
xh & 134 & 35.3 & 15.8 & 32.3 & 26.4 & 35.0 & - & - & - & \textbf{82.1} & \textbf{82.1}\\
xmf & 1389 & 73.4 & 85.0 & 77.9 & 78.7 & 77.7 & - & - & - & \textbf{87.9} & 87.7\\
yi & 2124 & 76.9 & 78.4 & 75.1 & 73.2 & 74.1 & - & - & - & 80.2 & \textbf{81.3}\\
yo & 3438 & 94.0 & 87.5 & 91.1 & 92.1 & 92.5 & 94.1 & 93.3 & 94.1 & 96.3 & \textbf{97.0}\\
za & 345 & 57.1 & 66.1 & 67.7 & 67.1 & 68.4 & - & - & - & 87.0 & \textbf{88.9}\\
zea & 7163 & 86.8 & 88.1 & 91.2 & 92.5 & 91.9 & - & - & - & 93.7 & \textbf{95.4}\\
zh & 1763819 & \textbf{82.0} & 78.7 & 78.6 & 80.4 & 78.2 & 77.2 & 78.5 & 79.2 & 58.3 & 76.6\\
zu & 425 & \textbf{82.3} & 61.5 & 61.0 & 70.7 & 70.3 & - & - & - & 79.6 & 80.4\\
\end{supertabular}